\documentclass{article}

\usepackage[final, main, nonatbib]{neurips_2025}

\usepackage[utf8]{inputenc} 
\usepackage[T1]{fontenc}    
\usepackage{hyperref}       
\usepackage{url}            
\usepackage{booktabs}       
\usepackage{amsfonts}       
\usepackage{nicefrac}       
\usepackage{microtype}      
\usepackage{xcolor}         
\usepackage{bibentry}
\usepackage{caption}
\usepackage{bm}
\usepackage{amsmath,amssymb}
\usepackage{bbm}
\usepackage{physics}  
\usepackage{enumitem}
\usepackage{graphicx}
\usepackage{subcaption}

\title{Mechanistic Interpretability of RNNs emulating Hidden Markov Models}

\author{
\textbf{Elia Torre}\thanks{\parbox[t]{0.9\linewidth}{Corresponding author: \texttt{name.surname@uzh.ch} \\ Code available at \url{https://github.com/EliaTorre/hmmrnn}}} \quad
\textbf{Michele Viscione} \quad
\textbf{Lucas Pompe} \quad
\textbf{Benjamin F. Grewe} \quad
\textbf{Valerio Mante} \\
\\
Institute of Neuroinformatics, University of Zurich \& ETH Zurich
}

\begin{document}
\maketitle

\begin{abstract}
Recurrent neural networks (RNNs) provide a powerful approach in neuroscience to infer latent dynamics in neural populations and to generate hypotheses about the neural computations underlying behavior. However, past work has focused on relatively simple, input-driven, and largely deterministic behaviors - little is known about the mechanisms that would allow RNNs to generate the richer, spontaneous, and potentially stochastic behaviors observed in natural settings. Modeling with Hidden Markov Models (HMMs) has revealed a segmentation of natural behaviors into discrete latent states with stochastic transitions between them, a type of dynamics that may appear at odds with the continuous state spaces implemented by RNNs. Here we first show that RNNs can replicate HMM emission statistics and then reverse-engineer the trained networks to uncover the mechanisms they implement. In the absence of inputs, the activity of trained RNNs collapses towards a single fixed point. When driven by stochastic input, trajectories instead exhibit noise-sustained dynamics along closed orbits. Rotation along these orbits modulates the emission probabilities and is governed by transitions between regions of slow, noise-driven dynamics connected by fast, deterministic transitions. The trained RNNs develop highly structured connectivity, with a small set of “kick neurons” initiating transitions between these regions. This mechanism emerges during training as the network shifts into a regime of stochastic resonance, enabling it to perform probabilistic computations. Analyses across multiple HMM architectures — fully connected, cyclic, and linear-chain — reveal that this solution generalizes through the modular reuse of the same dynamical motif, suggesting a compositional principle by which RNNs can emulate complex discrete latent dynamics.
\end{abstract}

\section{Introduction}
\label{sec:intro}
Modern large‑scale electrophysiology and imaging techniques provide access to the simultaneous activity of thousands of neurons in freely behaving animals, revealing the population‑level dynamics underlying perception, cognition, and movement \cite{spira2013multi, grienberger2012imaging, cunningham2014dimensionality, buonomano2009state, hausser2014optogenetics, steinmetz2021neuropixels}. In parallel to advances in neural recordings, it has become possible to obtain quantitative descriptions of unconstrained, natural behaviors across many timescales \cite{weinreb2024keypoint}. The combination of these techniques holds great promise in unlocking the neural computations underlying behavior \cite{datta2019computational}. 

Machine learning is critical to interpret and link the high-dimensional neural and behavioral data produced by modern neuroscience experiments. Hidden Markov Models (HMMs) can capture unconstrained, natural behaviors and decompose them into sequences of elemental motifs \cite{weinreb2024keypoint}. However, their discrete state representation may oversimplify the potentially continuous nature of neural processes. On the other hand, Recurrent Neural Networks (RNNs) provide a powerful approach to model the dynamics of large neural populations and to generate hypotheses about the underlying computations \cite{durstewitz2023reconstructing, zemlianova2024dynamical} . Despite this potential, RNNs are primarily used to model highly constrained tasks that are input-driven and largely deterministic. Hence, evidence for RNNs applicability in modelling stochastic discrete state transitions in a manner similar to HMMs remains limited. 

RNNs and HMMs may appear incompatible, as RNNs represent latent processes as trajectories through continuous state spaces, whereas HMMs rely on discrete latent states with stochastic transitions between them. Understanding whether RNNs can use continuous dynamics to generate stochastic transitions between discrete states would help bridge this conceptual gap and reduce the need for strong assumptions about the structure of the latent space. To address this question, we develop a training approach to directly fit RNNs to HMMs, and then reverse-engineer the trained RNNs to uncover the computations they implement. This strategy may ultimately lead to testable hypotheses about how biological neural circuits may implement discrete behavioral modes through continuous internal dynamics.

This work makes three main novel contributions:
\begin{enumerate}[leftmargin=*]
    \item A training paradigm for stochastic RNN behaviors (Sections~\ref{sec:setting}, Appendix~\ref{sec:training}, Figure~\ref{fig:pipe}): unlike previous work focused on deterministic tasks, we introduce a training method combining noise-driven RNNs with Sinkhorn divergence optimization, enabling learning of the stochastic state transitions typical of HMMs.
    \item A demonstration that RNNs can emulate HMM statistics (Sections~\ref{sec:setting}, Appendix~\ref{sec:metrics}, Figures~\ref{fig:euc},~\ref{fig:transition}): we show that vanilla RNNs accurately replicate the emission statistics of various HMM architectures, matching both transition dynamics and stationary distributions.
    \item A multi-level mechanistic account of how RNNs generate discrete stochastic outputs.
    \begin{enumerate}[label=3.\arabic*,leftmargin=2em]
        \item Global dynamics (Section~\ref{subsec:global}, Figures~\ref{fig:traj},~\ref{fig:cactus}): We reveal how training drives the RNNs into a regime where stochastic inputs sustain dynamics along closed orbits ("orbital dynamics"), while in their absence, the activity converges to a single fixed point.
        \item Local dynamics (Section~\ref{subsec:zones}, Figure~\ref{fig:zones}): We identify three functional zones: clusters, kick-zones, and transitions, each with distinct dynamical signatures.
        \item Connectivity and single neurons (Section~\ref{sec:causal}, Figures~\ref{fig:weights},~\ref{fig:interventions}): We uncover structured connectivity between "noise-integrating populations" and "kick neurons" and validate their role in triggering state transitions through causal interventions.
        \item Computational principle (Section~\ref{sec:discussion}, Figure~\ref{fig:oscillations}): We show the inferred mechanism relies on self-induced stochastic resonance.
    \end{enumerate}
\end{enumerate} 
Our findings form a coherent picture: local dynamical features, such as clusters and kick-zones, explain how global orbital dynamics give rise to discrete outputs, while the network’s connectivity reveals how individual neurons instantiate these dynamics. Together, these findings show that RNNs discover a general computational motif that combines slow noise integration with fast transitions. 

The mechanism we uncover offers a novel interpretation of how RNNs can generate stochastic transitions between discrete states. This mechanism (Contributions 3.1 - 3.4) acts as a “dynamical primitive”, reusable in a modular fashion to emulate more complex discrete latent structures. Multiple instances of this motif combine to govern transitions between specific pairs of states. 

Our results lay the foundation for future work, both in scaling to more complex HMM structures and in investigating whether similar dynamical motifs are employed by biological neural circuits during naturalistic behavior.

\section{Related Work}
\label{sec:related‑work}
RNNs have been trained on simple perceptual, cognitive, and movement tasks, both with supervised learning \cite{mante2013context, sussillo2015neural, wang2018flexible} and reinforcement learning \cite{wang2018prefrontal, yang2020artificial}. These RNNs are often trained with few or no constraints and then reverse-engineered to reveal potentially novel hypotheses about the computations performed by biological neural circuits. Fully understanding the dynamics of RNNs and characterizing the mechanisms they implement (i.e., mechanistic interpretability) remains a challenging and open problem \cite{smith2021reverse, sharkey2025open}. 

A common reverse-engineering approach focuses on characterizing the topology of fixed points and on obtaining linear approximations of the dynamics around them \cite{sussillo2013opening}. For example, RNNs trained on motion discrimination \cite{mante2013context} or on sentiment-classification \cite{maheswaranathan2019reverse} integrate inputs along line-attractors. Networks trained to reproduce reach movements implement rotational dynamics similar to that in primate motor cortex \cite{sussillo2015neural} and RNNs performing beat continuation generate low-dimensional oscillatory dynamics \cite{zemlianova2024dynamical}. Across many tasks, the topology of fixed points is invariant to the details of the RNN implementation (e.g. vanilla RNN, GRU, LSTM), suggesting a "universality" in the types of solutions implemented by RNNs \cite{maheswaranathan2019universality}. 

Recent work has pushed mechanistic intepretability beyond population dynamics, to the underlying RNN connectivity. Constraining RNNs weigths to being low-rank can reveal the relation between connectivity, computations, and dynamics \cite{mastrogiuseppe2018linking}. Such relations may extend to RNNs trained without constraints, as their function may rely on a low-rank component of the full connectivity \cite{krause2022operative, herbert2022impact}. Highly structured weights may emerge in particular when connectivity and latent dynamics are trained concurrently while imposing biological constraints \cite{langdon2025latent} or when training on many different tasks at once \cite{johnston2023abstract, driscoll2024flexible,riveland2024natural}.

Building on this line of work, here we investigate whether the above insights can be extended to RNNs implementing a different type of computation, namely internally-driven, probabilistic transitions between discrete latent states.

\section{Approximating HMMs with RNNs}
\label{sec:setting}
\begin{figure}[h]
    \centering
    \includegraphics[width=1.0\textwidth]{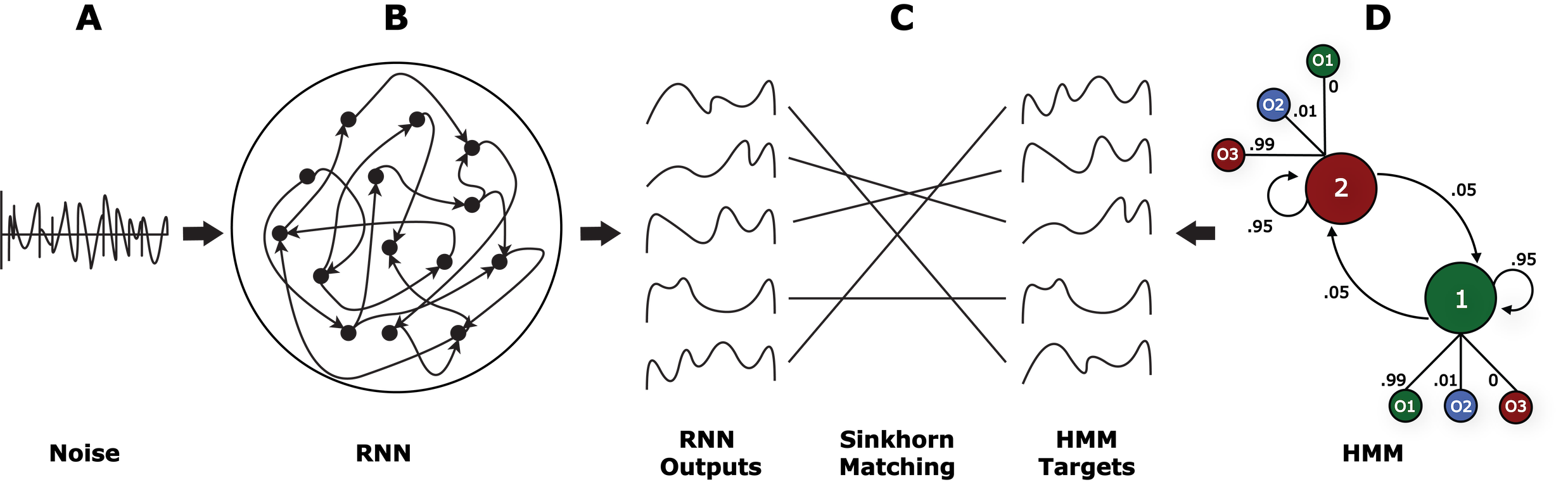}
    \caption{\textbf{Training pipeline.} 
        \textbf{A}: At each time‑step the network receives an i.i.d.\ Gaussian input vector
        \(x_t \sim \mathcal{N}(0, I_d)\), where the input dimensionality is varied across
        experiments (\(d \in \{1, 10, 100, 200\}\)).
        \textbf{B}: The signal is processed by a vanilla recurrent neural network,
        \(h_t = \text{ReLU}\!\bigl(h_{t-1} W_{hh}^{\mathsf T} + x_t W_{ih}^{\mathsf T}\bigr)\);
        the hidden‑state dimension is also varied
        (\(|h| \in \{50, 150, 200\}\)).
        Finally, the hidden state is mapped linearly to a three‑dimensional output via
        \(y_t = h_t A^{\mathsf T}\). To mimic HMM emissions, logits are converted to categorical samples via \textit{Gumbel–Softmax}. \textbf{C}: Parameters \(\Theta=\{W_{hh},W_{ih}, A\}\) are optimised by minimising the \emph{Sinkhorn divergence} between batches of predicted outputs \(Y=\{y_t\}_{t=1}^{T}\) and target sequences
        \(Y^{\star}=\{y^{\star}_s\}_{s=1}^{T}\). \textbf{D}: Target sequences \(Y^{\star}\) are generated by \textit{linear-chain}, \textit{fully-connected}, and \textit{cyclic} HMMs (described in Section~\ref{subsec:data‑gen} and Appendix \ref{sec:hmms}); the example shows a 2-state \textit{linear-chain} model.}
    \label{fig:pipe}
\end{figure}

To understand how RNNs encode discrete, probabilistic structure in their state spaces, we trained them to replicate the outputs of HMMs, which implement a process that is discrete and probabilistic by construction (Fig.~\ref{fig:pipe}). 

\subsection{Families of HMM architectures}
\label{subsec:data‑gen}
To investigate how RNNs encode discrete, probabilistic structure across different latent state topologies, we train them to replicate the outputs of HMMs with three distinct types of architectures: \textit{linear-chain}, \textit{fully-connected} and \textit{cyclic} structures.
\paragraph{Linear–chain HMMs.} We first consider a family of \textit{linear-chain} HMMs that span a spectrum from maximally discrete to quasi-continuous representations. These models have $M\!\in\!\{2,3,4,5\}$ latent states $\mathcal{S}^{(M)}=\{1,\dots,M\}$
and a common alphabet of observations $\mathcal{O}=\{1,2,3\}$.
Each model \(\mathcal{H}_M\) is fully specified by an emission matrix $\mathbf{E}^{(M)}\!=\![e^{(M)}_i]_{i=1}^{M}$ and a band‑diagonal
transition matrix $\mathbf{T}^{(M)}\!=\![t^{(M)}_{ij}]_{i,j=1}^{M}$.

For each latent state \(i\), we obtain emission probabilities as linear interpolation between a state skewed toward observation \(1\) (\(i=1\)) and one skewed toward observation \(3\) (\(i=M\)), while keeping $P(o_t=1)=\varepsilon$ constant. At each timestep, the chain either remains in the current state or transitions to a neighboring state. Given a change‑rate hyper‑parameter $\rho=0.05$, we set $q_M=\rho^{1/(M-1)}\in(0,\tfrac12)$, $\alpha_i=(i-1)/(M-1)$, fix $\varepsilon=0.01$ and define:

\[
e^{(M)}_i \;=\;
\begin{bmatrix}
(1-\varepsilon)\bigl(1-\alpha_i\bigr)\\[2pt]
\varepsilon\\[2pt]
(1-\varepsilon)\,\alpha_i
\end{bmatrix},
\quad
t^{(M)}_{ij}=
\begin{cases}
1-q_M,& i\in\{1,M\},\;j=i,\\
1-2q_M,& 1<i<M,\;j=i,\\
q_M,& j=i\pm1,\\
0,&\text{otherwise.}
\end{cases}
\tag{1}
\]
\noindent
This parametrization maintains a constant probability $\rho$ of reaching the most distant state in $M - 1$ steps. When \(M=2\) the system is maximally discrete; as \(M\) increases, the same $0\!\to\!2$ continuum is partitioned into progressively finer bins, yielding a quasi–continuous latent representation. This design systematically varies discreteness to reveal how RNN dynamics adapt as the latent space becomes more continuous.

\paragraph{Fully-connected HMMs.} To assess generalization beyond simple structures, we use a 3-state fully connected HMM where each state can transition to any other. Each state favors one of the three outputs but retains small probabilities for the others, testing RNNs on fully interconnected dynamics.

\paragraph{Cyclic HMMs.} We also consider a 4-state cyclic HMM with bidirectional closed-loop transitions. For each state, emissions are biased toward a characteristic pair of outputs, one dominant and one weaker component, with adjacent states sharing a common output.

The complete specification of all HMM architectures, including transition and emission matrices, is provided in Appendix~\ref{sec:hmms} (Fig.~\ref{fig:hmms}). 

\subsection{Training Networks with Sinkhorn Loss Function and Performance Metrics}
\label{subsec:architecture}
\paragraph{Network Architecture.} We employ standard, \textit{vanilla} RNNs \cite{elman1990finding} of hidden-state size $|h| \in \{50, 150, 200\}$. At each time-step, the network receives Gaussian input \(x_t \sim \mathcal{N}(0, I_d)\), with \(d \in \{1, 10, 100, 200\}\). The hidden state is updated and projects onto the three logits via: 
\[
    h_t = \text{ReLU}(h_{t-1} W_{hh}^{\mathsf T} + x_t W_{ih}^{\mathsf T}), \quad y_t = h_t A^{\mathsf T}.
    \tag{2}
\]

To mimic the discrete emissions of an HMM, these logits are converted to categorical samples using the \textit{Gumbel-Softmax reparametrization trick}: we add i.i.d. Gumbel noise, divide by a temperature $\tau$ (set to 1 in all experiments), and apply a soft‑max. This continuous relaxation acts as a differentiable proxy for the non‑differentiable \textit{argmax} in the Gumbel‑Max method, letting gradients flow through the sampling step while still converging to one‑hot vectors as $\tau \rightarrow 0$ \cite{jang2016categorical} \cite{maddison2016concrete}.

\paragraph{Sinkhorn Divergence.} Because our target sequences are probabilistic, we use a loss function that is appropriate for comparing distributions: the \emph{Sinkhorn divergence} \cite{feydy2019interpolating}\cite{genevay2018learning}\cite{sourmpis2023trial}, an optimal transport (OT) distance. Typically, OT-distances find a binary coupling matrix $\Pi$ linking single samples in the outputs and targets, which minimizes the total Euclidean distance between coupled samples. If the euclidean distance between coupled samples is zero, the output distribution reproduces the target distribution. Finding such a coupling matrix is computationally expensive and non-differentiable. The Sinkhorn divergence overcomes these issues by finding a smoothed coupling matrix allowing non-unique couplings \cite{cuturi2013}. Details on the training regime are provided in Appendix~\ref{sec:training}.

\paragraph{Performance Metrics.} To quantify how closely the trained RNNs replicate their target HMMs, we track four statistics:
\textbf{(i)} Euclidean distance between "Sinkhorn‑aligned" output sequences (global reconstruction error);
\textbf{(ii)} the $3 \times 3$ transition matrix of successive emissions (long‑range dynamics);
\textbf{(iii)} marginal observation frequencies (stationary distribution); and
\textbf{(iv)} observation volatility (proportion of time steps with output changes). For all metrics, the trained RNNs replicate the emission statistics of the target HMMs. Definitions and results for all combinations of hidden size ($|h| \in {50, 150, 200}$) and input dimensionality ($d \in {1, 10, 100, 200}$) are shown in Appendix~\ref{sec:metrics} (Figs.~\ref{fig:euc}, \ref{fig:transition}). Higher input dimensionality improves convergence, while increasing hidden size beyond 150 yields only marginal benefits. Below, we thus focus on the four configurations with $|h| = 150$ and $d = 100$.

\section{Mechanistic Interpretability: Latent Dynamics}
\label{sec:black‑box}

\begin{figure}[h]
    \centering
    \includegraphics[width=1.0\textwidth]{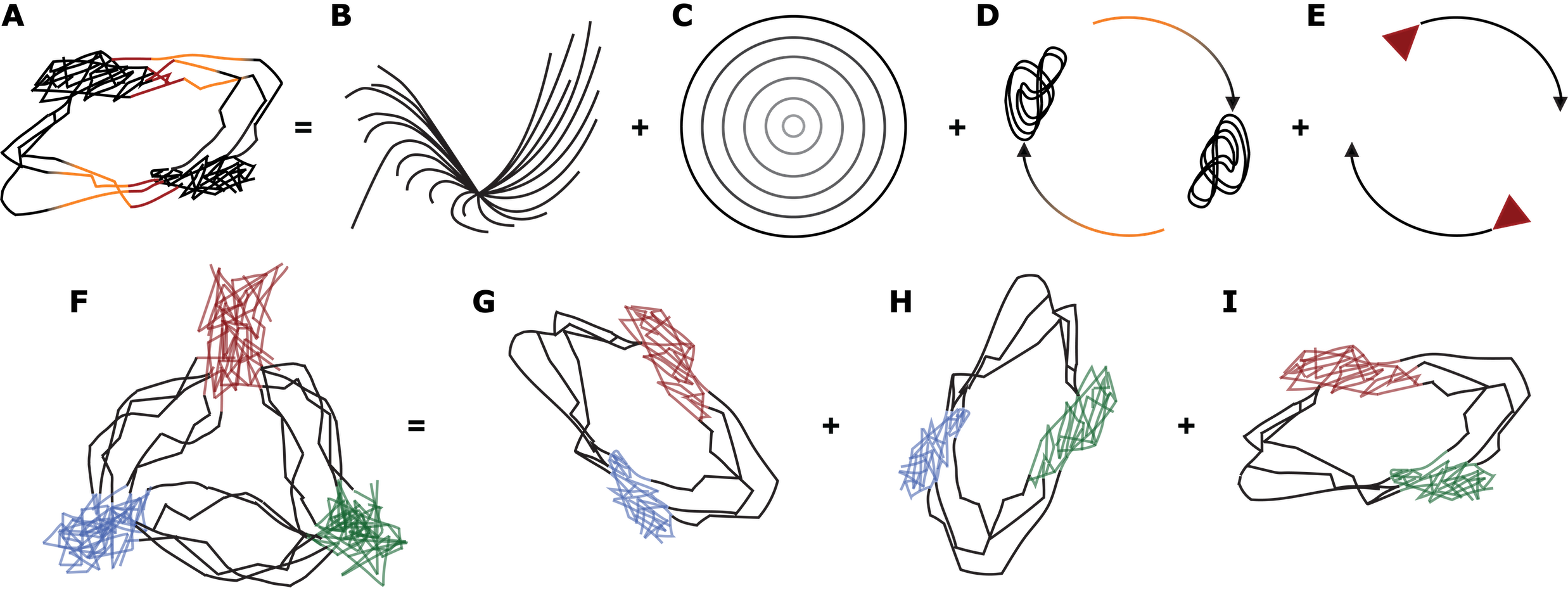}
    \caption{\textbf{A compositional dynamical primitive}. Schematic of the dynamical motifs uncovered through reverse-engineering of trained RNNs. \textbf{A}: Sample trajectory in the continuous state space. \textbf{B}: A single fixed point in the absence of input. \textbf{C}: Stochastic input displaces the system from this attractor, sustaining dynamics along closed orbits whose radius grows with input variance. \textbf{D}: Distinct slow regions (\textit{clusters}) where trajectories linger, separated by \textit{transition} regions. \textbf{E}: Local “\textit{kick‑zones}” that trigger transitions between clusters. \textbf{F-I}: Schematic of a richer discrete latent structure (\textbf{F}) expressed as a composition of three instances of the same dynamical primitive (\textbf{G-I}). Each instance preserves the same global and local dynamical properties described in panels (\textbf{A-E}).}
    \label{fig:sketch}
\end{figure}

To uncover the mechanisms implemented by the trained RNNs, we first consider their global (Section~\ref{subsec:global}) and local (\ref{subsec:zones}) latent dynamics. These analyses reveal how several distinct "dynamical motifs" together contribute to the RNNs' ability to approximate HMM-like emission statistics (Fig.~\ref{fig:sketch}).

\subsection{Global Latent Dynamics: Noise-sustained Orbital Dynamics}
\label{subsec:global}
\begin{figure}[h]
    \centering
    \includegraphics[width=1.0\textwidth]{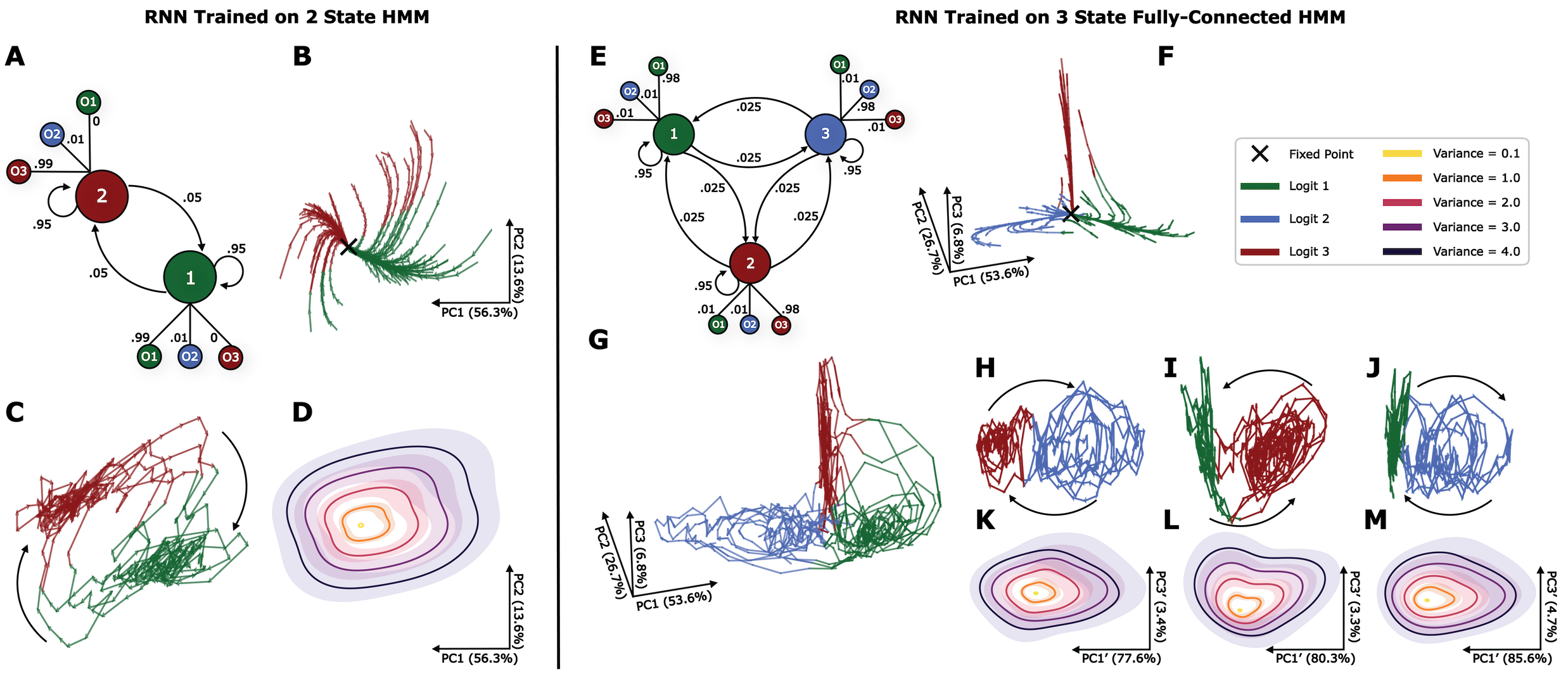}
    \caption{\textbf{Global latent dynamics of trained RNNs.} Panels \textbf{A} and \textbf{E} illustrate the HMM architectures approximated by the corresponding RNNs shown on the left and right. Hidden-state trajectories are projected onto the first two (or three) principal components of the activity and colored by the dominant logit. \textbf{B}, \textbf{F}: in the absence of input, trajectories from random initial conditions converge to a single fixed point (cross). \textbf{C}, \textbf{G}: under Gaussian input noise, activity evolves along stable orbits with slow regions associated to distinct outputs; arrows indicate flow. More complex dynamics in \textbf{G} can be decomposed into the same fundamental motif observed in \textbf{C} by applying a second-level PCA within the first PCA space, shown in \textbf{H–J}. \textbf{D}, \textbf{K–M}: contour plots (95\% CI) of hidden-state densities under increasing input variance (\(\sigma^2 \in \{0.1, 1.0, 2.0, 3.0, 4.0\}\)) reveal a linear scaling between input variance and orbit radius. Results for all HMM structures are provided in Appendix Figures~\ref{fig:trajectories_linearchain}, \ref{fig:trajectories_fullyconnected}.}
    \label{fig:traj}
\end{figure}

\paragraph{Orbital dynamics under stochastic input.} We investigate the global latent dynamics of the trained RNNs by projecting their hidden states onto the first two principal components (PCA, Figure~\ref{fig:traj}). When initialized from random hidden states, and without external input, activity converged to a single fixed point (Figs.~\ref{fig:sketch}B, \ref{fig:traj}B), with no evidence for separate attractors that might be expected to represent the discrete latent states of the target HMMs. However, the latent dynamic changes markedly for RNNs receiving stochastic (Gaussian) inputs, to a regime where trajectories exhibit orbital dynamics (Figs.~\ref{fig:sketch}A, \ref{fig:traj}C). The stochastic input pushes activity outward (Figs.~\ref{fig:sketch}C, \ref{fig:traj}D), while the recurrent component pulls it back, together implementing a stable closed orbit along which activity evolves uni-directionally. Along this orbit, the RNN exhibits regions of slow dynamics (\textit{clusters}) each corresponding to a distinct output class of the reference HMMs, with \textit{transitions} occurring between them (Fig.~\ref{fig:sketch}D). Critically, transitions between slow zones result in large changes in the output probabilities. As the number of latent states increases in \textit{linear-chain} HMMs, RNNs do not form additional slow regions along the orbits (App. Fig.~\ref{fig:trajectories_linearchain}). Instead, they capture finer emission discretizations by modulating the alignment of readout axes with the plane containing the orbital dynamics (Appendix~\ref{sec:alignment}, Fig.\ref{fig:alignment}). In contrast, for \textit{fully-connected} and \textit{cyclic} architectures, RNNs develop multiple orbits connecting distinct pairs of slow regions, reflecting the richer structure of the underlying HMMs. These dynamics reveal a \textit{compositional dynamical motif} in which the same fundamental unit combines to generate increasingly complex latent structures (Fig.~\ref{fig:traj}G and App. Fig.~\ref{fig:trajectories_fullyconnected}).

\begin{figure}[h]
    \centering
    \includegraphics[width=1.0\textwidth]{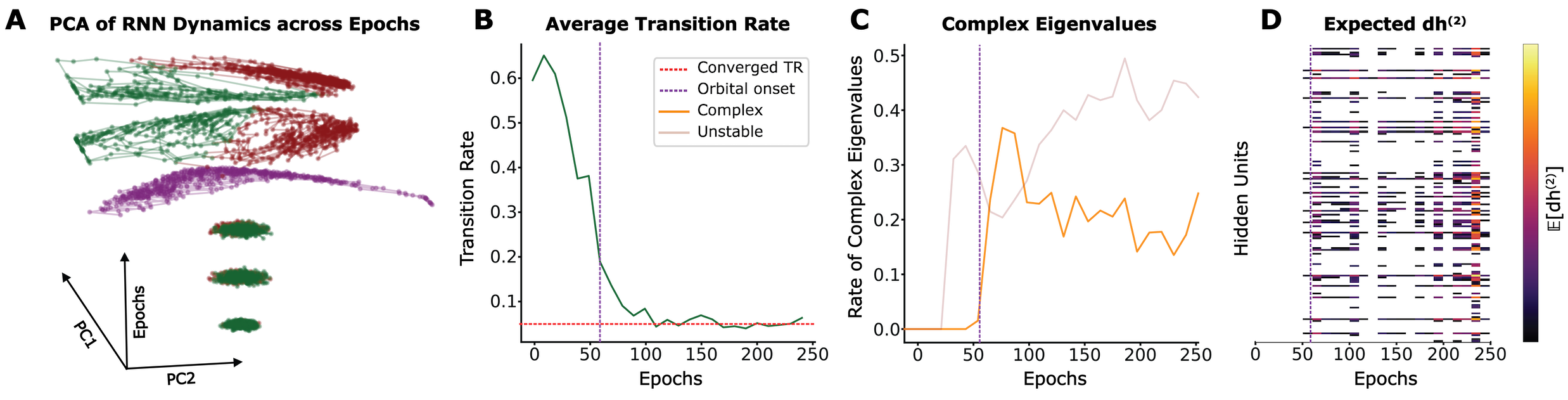}
    \caption{\textbf{A:} PCA projections of the latent dynamics across epochs for an RNN trained on a 2-states HMM, green and red indicate the dominant logit across epochs. The model first learns a single-fixed point, then becomes unstable (purple), and finally transitions to orbital dynamics. \textbf{B:} The RNN learns to encode emission probabilities of the target HMM, by means of a specific rate of transitions between the two \textit{clusters}. \textbf{C:} The RNN becomes unstable just before the transition, then the complex eigenvalues appear on the imaginary axis. \textbf{D:} The expected second order perturbation vector $\mathbb{E}[dh^{(2)}]$ emerges in the proximity of the transition. This perturbation approximates how the variance of the noise affects the free recurrent dynamics, capturing the dependency of the orbital dynamics on the input noise variance. Results validating these findings for all HMM structures are provided in Appendix~\ref{sec:learning_traj} (Figs.~\ref{fig:learning_traj}, \ref{fig:learning_traj2}).}
    \label{fig:cactus}
\end{figure}

\paragraph{Emergence of orbital dynamics during training.} PCA projections reveal a clear transition in the latent dynamics across training epochs (Fig.~\ref{fig:cactus}A, purple), marking the shift from a stable fixed point to orbital dynamics. As training progresses, unstable eigenvalues emerge, marking the destabilization of the fixed-point regime. This transition coincides with a rise in eigenvalues near the imaginary axis, indicating the onset of oscillatory activity (Fig.\ref{fig:cactus}C), and with a characteristic \textit{double-descent} in the loss curve \cite{eisenmann2023bifurcations} (Fig.\ref{fig:loss}). Functionally, this transition enables sustained quasi-periodic oscillations that encode the temporal dynamics of the target HMM (Fig.~\ref{fig:oscillations}). The radius of the resulting orbits scales linearly with input variance (Fig.~\ref{fig:traj}D), a relationship confirmed by a perturbative analysis of the RNN dynamics (Appendix~\ref{sec:second order perturbation}). Under unbiased Gaussian input, first-order perturbations average out, while second-order terms —scaling linearly with variance — dominate after the transition, explaining how stochastic input shapes the emergent orbital regime (Fig.\ref{fig:cactus}D). Over training, the network converges to a stable average rate of transitions between clusters (Fig.\ref{fig:cactus}B). In Section~\ref{sec:causal}, we examine how this behavior arises from the interplay between recurrent connectivity and input noise.

\subsection{Local Latent Dynamics: Clusters, Transitions, and Kick‑Zones}
\label{subsec:zones}
\begin{figure}[h]
    \centering
    \includegraphics[width=1.0\textwidth]{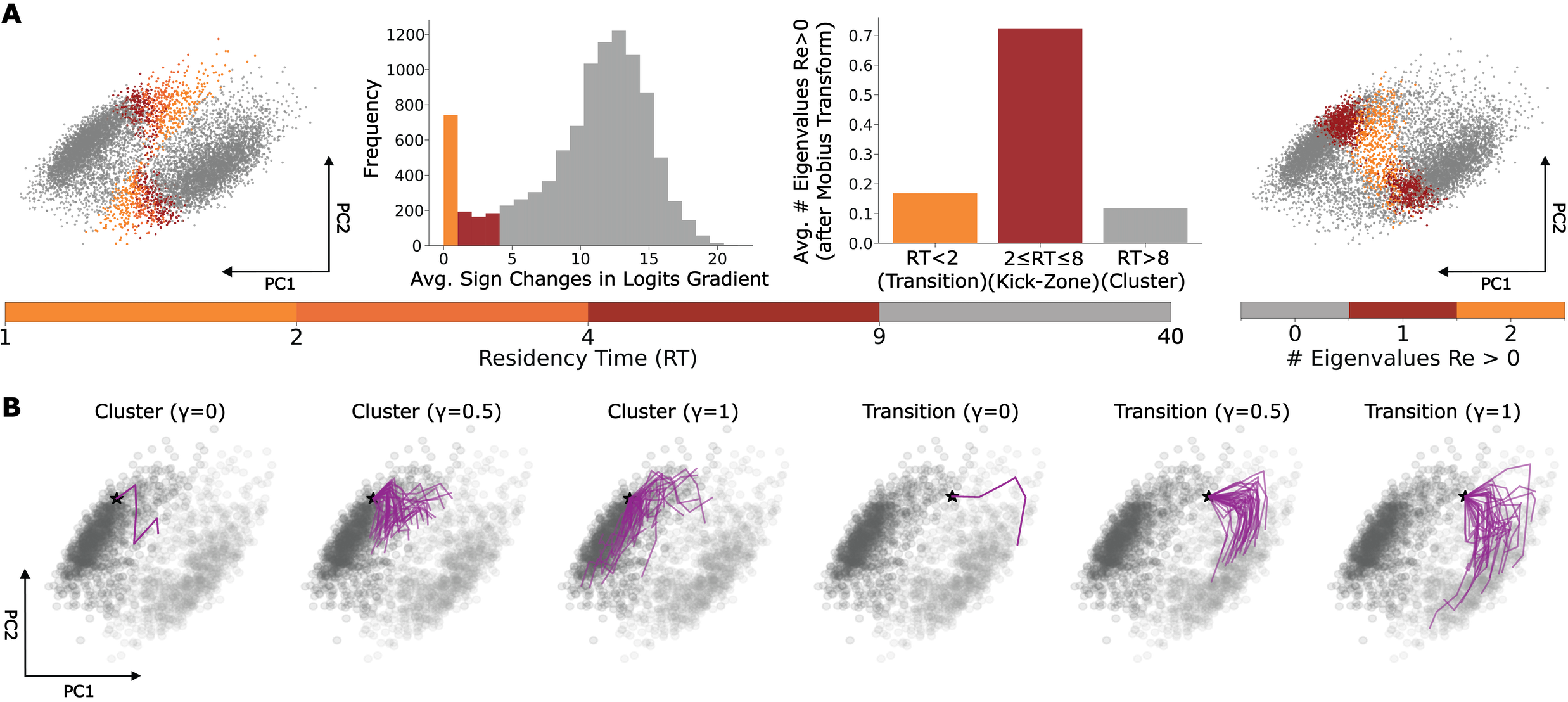}
    \caption{\textbf{Clusters, Transitions, and Kick-Zones.} 
    \textbf{A}: Left to right: (i) Residency time (RT) reveals slow (gray) and fast (colored) regions; (ii) logit-gradient sign changes histogram shows a bi-modal distribution, separating stable clusters from transitions; (iii) average number of unstable directions per region (via Möbius-transformed Jacobian), peaks in kick-zones and is lowest in clusters; (iv) spatial distribution of the number of real eigenvalues with positive real part, highlighting localized instability in kick-zones contrasted with the stability of clusters. \textbf{B}: Noise sensitivity analysis from sampled states (black star) in clusters (left) and transitions (right) with 30 trajectories (in purple) generated under increasing noise resampling conditions (\(\gamma \in \{0, 0.5, 1\}\)). Transitions are robust across noise conditions, while clusters exhibit increasing dispersion, indicating higher noise sensitivity. Panels \textbf{A}, \textbf{B} show results for RNNs trained on 5-state HMMs. Results validating these findings for the remaining configurations are provided in Appendix~\ref{sec:residency_graph} (Figs.~\ref{fig:residency_graph1},\ref{fig:residency_graph2},\ref{fig:residency_graph3} and \ref{fig:sensitivity}).}
    \label{fig:zones}
\end{figure}

Having characterized the global dynamics, we zoom in on the local properties of the orbital dynamics to understand \emph{where} and \emph{how} the network implements transitions between slow zones. Standard linearization techniques around fixed points (Section~\ref{sec:related‑work}) offer limited insight here, as the observed convergence towards the single fixed point does not explain the full emergent dynamics (Fig.~\ref{fig:traj}). More insights can be obtained by considering how activity evolves in the presence of the input noise on different, short rollouts initialized from a given state-space location (Fig.~\ref{fig:zones}). Specifically, we segment state-space into different "zones" based on three measures computed at each location: the \emph{residency time} (RT), defined as the average number of steps required in a rollout until a change in the dominant logit occurs; the \emph{logit sign-change count}, quantifying how often the dominant logit's gradient flips before a change in the dominant logit occurs, with lower values indicating more directed flow; and the \emph{number of unstable directions}, obtained via Jacobian linearization and Möbius-transformed eigenvalues, which reveals local sensitivity to perturbations (Appendix~\ref{app:jacobian}).

\paragraph{Three dynamical regimes.} Segmenting state space by residency time (RT) reveals three functional zones, each with distinct dynamical signatures (Fig.~\ref{fig:zones}A). We find similar zones in all RNNs, irrespective of the architecture of the target HMM (Appendix~\ref{sec:residency_graph}: Figs.~\ref{fig:residency_graph1},\ref{fig:residency_graph2},\ref{fig:residency_graph3}).

\begin{itemize}[left=0.5em, labelwidth=2.5cm, labelsep=0.5em, align=left]
  \setlength\itemsep{4pt}
  \item[\textbf{Clusters}] (RT \(>\) 8): regions where trajectories linger the longest, with frequent logit‑gradient sign changes (in the range 5–20), and essentially only contracting eigenvalues. These are locally stable regions, each corresponding to a different probability distribution over the outputs.

  \item[\textbf{Kick‑zones}] (2 \(\leq\) RT \(\leq\) 8): located downstream of clusters, these regions exhibit moderate logit‑gradient sign changes (around 2–4) and a few unstable directions locally stretch the flow, indicating a local push away from the cluster.

  \item[\textbf{Transitions}] (RT \(<\) 2): Once trajectories cross the kick‑zone, they enter short-lived corridors where the system moves nearly deterministically toward the next cluster. These regions exhibit few logit‑gradient sign changes ($< 1$), reflecting a stable and directed flow. 
\end{itemize}

\paragraph{Noise Sensitivity.}  
To further validate the functional relevance of these regions we explicitly probe their sensitivity to noise (Fig.~\ref{fig:zones}B) for the RNN trained on a 2-state HMM; Appendix~\ref{sec:residency_graph}, Fig.~\ref{fig:sensitivity} for all other \textit{linear-chain} RNNs. We sample initial conditions from both \textit{cluster} and \textit{transition} regions and generate trajectories under three noise conditions: \textit{identical} (\(\gamma = 0\)), \textit{partially resampled} (\(\gamma = 0.5\)), and \textit{fully independent} (\(\gamma = 1\)). Transition regions show minimal variability across noise conditions — once the kick-zone is crossed, trajectories proceed quasi-deterministically toward the next cluster. In contrast, cluster regions are noise-sensitive: as \(\gamma\) increases, trajectories diverge, with some crossing the kick-zone and others returning to the cluster. We validated these qualitative differences with quantitative measures of divergence: (i) the trace of the covariance matrix per timestep, capturing dispersion; and (ii) the time-course of the average Euclidean distance between individual trajectories and the mean trajectory (Appendix~\ref{sec:residency_graph}, Fig.~\ref{fig:trace}).

\section{Mechanistic Interpretability: single neuron computations and connectivity}
\label{sec:causal}
In the previous section we described a computational mechanisms at the level of population-level dynamics, which relies in particular on transitions between slow regions. Here we aim to explain how these transitions emerge based on key features of the RNN connectivity and the resulting single unit properties. We focus on an RNN trained on the two-state HMM and report the analyses for the other configurations in Appendices~\ref{sec:residency_graph}, \ref{sec:neurons} and \ref{sec:circuit}.

\paragraph{Discovering “kick‑neurons”.}
Among all RNN units, \textit{two separate triplets of neurons} have pre-activation values (before \textit{ReLU}) with a distinctive spatial profile: pre-activation values are strongly negative within \textit{clusters}, pass through a near-zero regime in the \textit{kick‑zones}, and become positive during \textit{transitions} (Appendix~\ref{sec:neurons}, Fig.~\ref{fig:neurons}). The intermediate regime places the units near the \textit{ReLU} activation threshold, where small variations in input can determine the opening of the \textit{ReLU gate} — a mechanism we describe in the next paragraph. Each triplet is linked to one transition direction, firing at the onset of movement between clusters and emerging as the dominant non-zero components of the second-order perturbation vector $\mathbb{E}[dh^{(2)}]$ (described in Section~\ref{subsec:global}), which reflects the network’s sensitivity to input variance. Together, these properties suggest a causal role in generating noise-driven "kicks" that initiate state changes, prompting us to term them \emph{kick-neurons}.

\begin{figure}[h]
    \centering
    \includegraphics[width=1.0\textwidth]{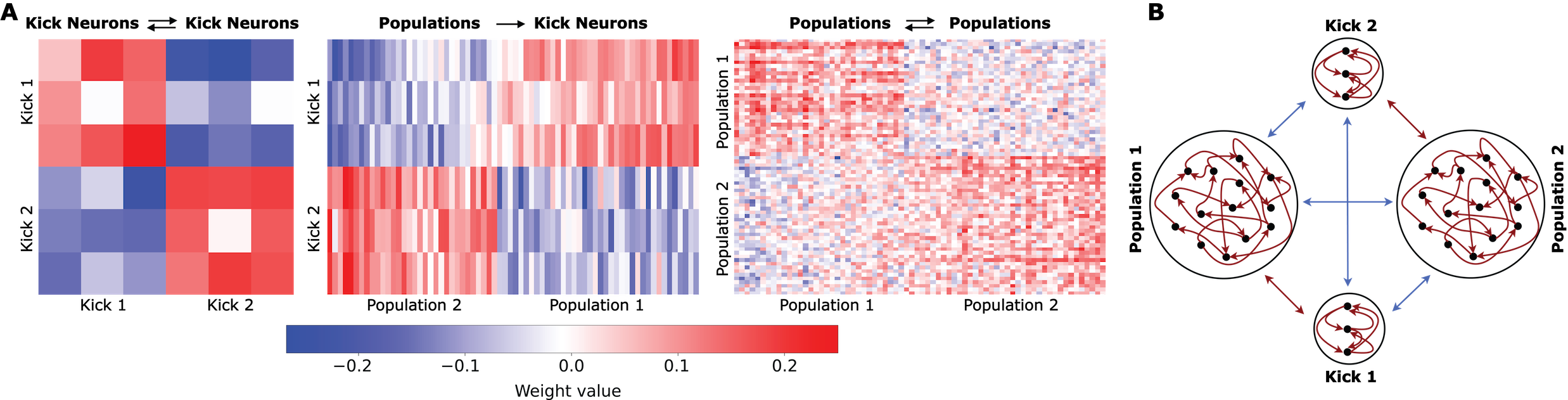}
    \caption{\textbf{Recurrent weight matrix and \textit{Kick-circuit}.} \textbf{A}: Left to right: (i) Sub‑matrix of $W_{hh}$ restricted to the six \textit{kick‑neurons}. Within‑triplet weights are positive (red), whereas cross‑triplet weights are negative (blue), indicating mutual excitation and reciprocal inhibition. (ii) Weights from \textit{noise‑integrating neurons} to \textit{kick‑neurons} (sorted). Each integrating population excites one triplet and inhibits the other. (iii) Recurrent weights within the integrating populations show within‑population excitation and cross‑population inhibition. \textbf{B}: \textit{Kick-circuit}: red arrows indicate excitation, blue inhibition. The circuit comprises two self-exciting, mutually inhibiting loops that project to opposing \textit{kick-neuron} triplets, forming a noise-integration mechanism that implements stochastic transitions between slow regions. Results for linear-chain and fully-connected HMMs in App. Figs.~\ref{fig:circuit1}, \ref{fig:circuit2}.}
    \label{fig:weights}
\end{figure}

\paragraph{Noise as the trigger.}
To better understand the \textit{kick-mechanism}, we examine the recurrent weight matrix ($W_{hh}$) and find that \emph{kick-neurons} within each triplet form mutually excitatory connections, while inhibiting the opposing triplet (Fig.~\ref{fig:weights}). Similar to work by \cite{fanthomme2021low}, showing that input noise integration in RNNs can give rise to independent subpopulations, we observe two larger neuronal groups (comprising $\sim 70$ neurons), each forming recurrent excitatory loops within themselves while projecting inhibitory connections to the other. These populations interact with the \emph{kick-neurons} through structured excitatory and inhibitory connections, suggesting a role as noise integrators that modulate the \textit{transition gate} opening, hence we refer to them as \emph{noise-integrating populations}.

\begin{figure}[h]
    \centering
    \includegraphics[width=1.0\textwidth]{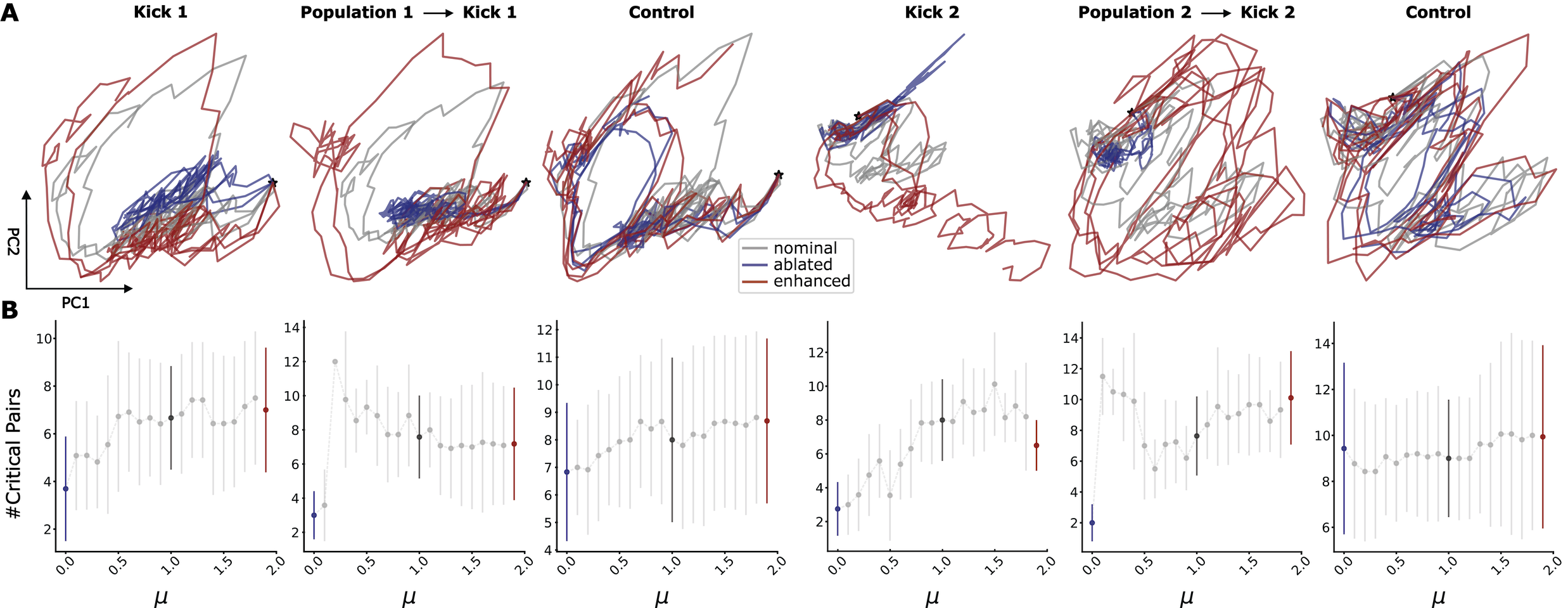}
    \caption{\textbf{Causal interventions validate the kick-circuit.} \textbf{A}: Latent trajectories in PCA projection. We sample initial conditions (ICs) from the two cluster regions and let trajectories evolve from the same IC with identical input noise draws in unperturbed (gray), ablated (blue), and enhanced (red) regimes for 60 time-steps. Suppressing either the \textit{kick‑neurons} (columns: 1, 4) or their noise-drive from \textit{integrating populations} (columns: 2, 5) prevents transitions between clusters, while enhancement overshoots trajectories past the opposite cluster. Control manipulations on neurons not part of \textit{noise-integrating populations} (columns: 3, 6) preserve \textit{kick‑neurons} noise-drive and "cluster-switching". \textbf{B}: Stability signature of the interventions. Mean±s.d. number of critical eigenvalue pairs as a function of the modulation factor $\mu$. Ablation ($\mu = 0$, blue) eliminates \textit{critical-pairs}, consistent with reversion to a single fixed-point regime; enhancement ($\mu = 2$, red) maintains the number of \textit{critical-pairs}, preserving the orbital dynamics. Grey points show intermediate values of $\mu$.}
    \label{fig:interventions}
\end{figure}

\paragraph{Causal Interventions.} To confirm this mechanism, we performed targeted interventions (Figure~\ref{fig:interventions}). We modulated either the neurons' activity directly, or the noise-drive they receive through projections from the associated \textit{noise-integrating populations}. A modulation factor $\mu$ controlled both types of perturbations.

\begin{itemize}[left=0.5em, labelwidth=2.5cm, labelsep=0.5em, align=left]
\setlength\itemsep{4pt}
\item[\textbf{Ablation} ($\mu=0$).] Silencing the \textit{kick-neurons} or ablating the input noise to the corresponding \textit{integrating population} traps the trajectory within its current cluster, preventing state transitions. Both manipulations mirror each other, confirming that noise-driven activation is required to open the \textit{transition gate}. This intervention leads to a loss of critical eigenvalue pairs from the Jacobian spectrum, reflecting the collapse of the orbital dynamics and reversion to a single fixed-point regime.

\item[\textbf{Enhancement} ($\mu=2$).] Doubling \textit{kick-neuron} activity or amplifying the noise-drive from the corresponding \textit{integrating population} causes overshoots beyond the target cluster. These effects resemble an increase in noise variance, and the number of critical eigenvalue pairs remains unchanged, indicating that enhancing activity preserves the rotational regime in this setting.
\end{itemize}

\section{Stochastic Resonance}
\label{sec:discussion}
\begin{figure}[h]
    \centering
    \includegraphics[width=1.0\textwidth]{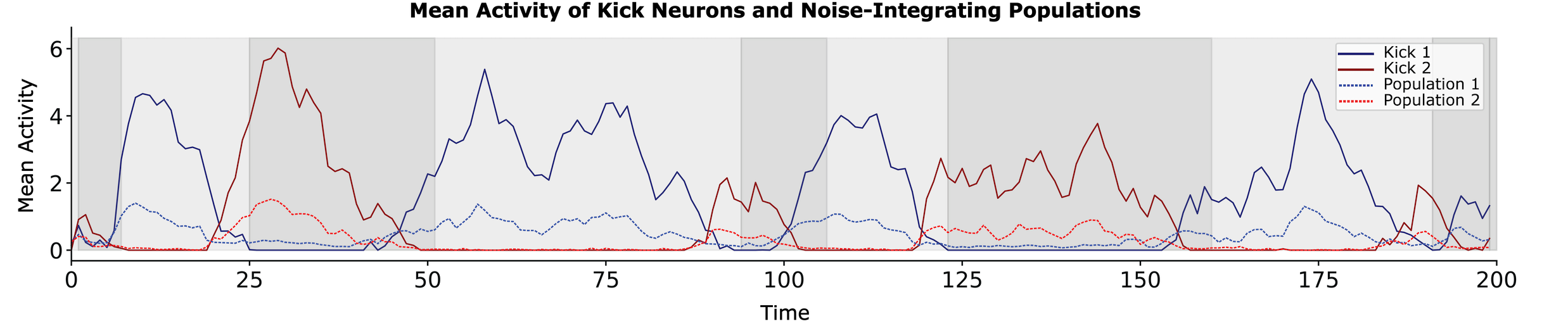}
    \caption{\textbf{Oscillatory dynamics.} Mean activity of kick-neuron triplets (solid lines) and noise-integrating populations (dashed lines) as the RNN switches cluster regions (alternating gray bands). These populations exhibit clear anti-phase oscillations: while one integrating group dominates, it reinforces itself and its matching kick-neurons, driving the system toward a transition. This dynamic reflects a \textit{self-induced stochastic resonance} (SISR) regime \cite{muratov2005self, zhu2021stochastic}, where slow noise integration and fast resets ("kicks") produce quasi-periodic switching without external periodic input.}
    \label{fig:oscillations}
\end{figure}

The above analyses show that our RNN exhibits noise-sustained orbital dynamics (Figure~\ref{fig:traj}), driven by structured interactions between two functionally distinct neural populations: slow, \textit{noise-integrating} units, and fast-responding \textit{kick-neurons} (Figure~\ref{fig:zones}, \ref{fig:weights}). As noise accumulates in the slow subsystem, trajectories drift along quasi-stable manifolds until they reach a region — the \textit{kick-zone} — which triggers a rapid, deterministic transition to the next attractor-like cluster. This mechanism results in robust, quasi-periodic alternation between network states, even in the absence of any external periodic input. This phenomenon bears resemblance to a class of noise-induced dynamics known as \emph{self-induced stochastic resonance} (SISR) \cite{muratov2005self}. In contrast to classical stochastic resonance, which requires an external periodic signal \cite{benzi1981mechanism}, SISR arises intrinsically in systems with time-scale separation. Weak noise applied to a fast subsystem perturbs the system off a slow manifold, initiating deterministic excursions that form coherent, noise-controlled oscillations \cite{zhu2021stochastic}. Our RNN appears to operate in an analogous regime: the slow populations accumulate stochastic input over time while in the cluster zones, whereas the fast \textit{kick-neurons} drive sharp transitions once a noise-modulated threshold is reached. This mechanism enables the emergence of stable oscillatory patterns, with their period governed by the interplay between noise variance and slow integration dynamics. The oscillation period, linked in our case to the emission probabilities, can be shaped through learning by adjusting the effective time-constant of the slow subsystem. This interpretation is supported by the population-level dynamics shown in Figure~\ref{fig:oscillations}, where each cycle reflects a transition between attractor-like states, coordinated through the interaction of slow integrators and fast responders whose activity rises sharply before the transition. In this way, the network effectively harnesses internal noise as a computational signal, leveraging SISR-like dynamics to perform structured, probabilistic inference, thus allowing the RNN to emulate the HMM behavior.

\section{Conclusion}
\label{sec:conclusion}
In this work, we explored how Recurrent Neural Networks (RNNs) can implement discrete, stochastic latent structure through continuous dynamics. We introduced a training pipeline that fits \emph{vanilla} RNNs to various families of Hidden Markov Models (HMMs). Contrary to the expectation of an $n$-well landscape with one fixed point per HMM state, trained RNNs converge to a single fixed point in the absence of input and, under noise, exhibit noise-sustained orbital dynamics: slow regions that encode distinct emission probabilities, separated by short, deterministic transitions. Mechanistically, it relies on two complementary sub-circuits: large noise-integrating populations and fast kick-neurons, whose interplay converts input variance into quasi-periodic transitions. This connectivity structure harnesses internal noise as a computational signal, enabling RNNs to reproduce HMM-like probabilistic behavior via a reusable dynamical motif and paralleling features of neural activity observed in the brain.

Cortical activity is intrinsically noisy at both micro- and macro-scales, driven by stochastic ion channel dynamics \cite{dangerfield2010stochastic}, probabilistic synaptic transmission \cite{allen1994evaluation}, and the high-dimensional, recurrent nature of cortical circuits, which produces ongoing, noisy background activity even in the absence of external input \cite{faisal2008noise}, potentially as a consequence of chaotic dynamics \cite{depasquale2023centrality}. Far from being a nuisance, such variability can enhance information processing — increasing sensitivity to weak or sub-threshold signals and facilitating the coordination of distributed brain regions in the processing of sensory information via \textit{stochastic resonance} \cite{gluckman1996stochastic, mcdonnell2009stochastic, pisarchik2023coherence}.

Our findings demonstrate that unconstrained RNNs can uncover both discrete and continuous latent structure directly from data — without any imposed topological priors — and reveal that these networks naturally converge toward biologically plausible circuit motifs. Taken together, these results point to a compositional dynamical primitive in which slow noise integration and fast kick-triggered resets cooperate to generate discrete state transitions, and multiple instances of this motif can combine to produce richer latent dynamics. Overall, this work positions Recurrent Neural Networks as a powerful alternative to Hidden Markov Models for modeling the latent structure and neural mechanisms of natural behaviors.

\clearpage
\bibliography{refs}

\clearpage
\appendix
\addcontentsline{toc}{section}{Appendix}
\begingroup
\setlength{\parskip}{0.1em}
\setlength{\itemsep}{0.1em}
\begin{small}
\tableofcontents
\end{small}
\endgroup

\clearpage
\section{Hidden Markov Models Architectures}
\label{sec:hmms}
\begin{figure}[h]
    \centering
    \includegraphics[width=1.0\textwidth]{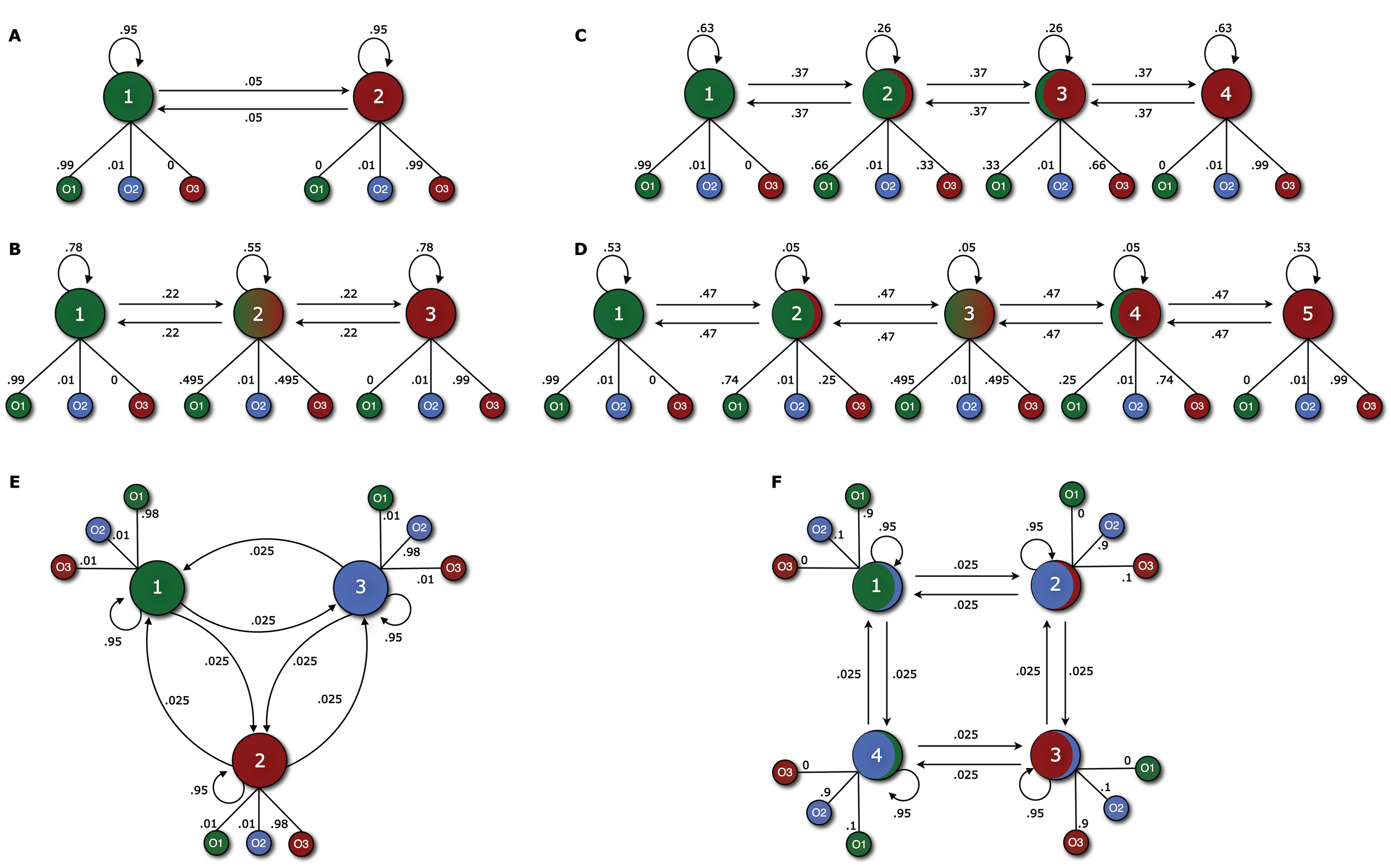}
    \caption{\textbf{Hidden Markov Models architectures used for training.} Panels \textbf{A-D} illustrate a spectrum of \textit{linear-chain} HMM with \(M\in\{2,3,4,5\}\) latent states (from left to right, top to bottom). Arrows indicate state transitions, with labels showing transition probabilities;  The hidden states of the HMM (larger circles) have been colored according to the most likely outcome associated with a specific state, or by a mixture of them. Smaller circles represent emission distributions for each state over the observation alphabet \(\mathcal{O} = \{1, 2, 3\}\), with green, blue, and red corresponding to emissions \(o_1\), \(o_2\), and \(o_3\), respectively. All models are constructed to preserve a constant probability \(\rho = 0.05\) of reaching the most distant state in \(M{-}1\) steps. As \(M\) increases, the latent representation becomes progressively finer, transitioning from highly discrete (\textbf{A}) to quasi-continuous (\textbf{D}). \textbf{E}: \textit{Fully-connected} HMM with three latent states, each capable of transitioning to any other (including self-transitions). Emission probabilities are biased toward one dominant output per state, while retaining smaller weights for the others, producing a symmetric, all-to-all transition structure. \textbf{F}: \textit{Cyclic} HMM with four latent states arranged in a bidirectional loop. Each state emits a characteristic pair of outputs, with one output dominant and the other weaker, while adjacent states share one common output.}
    \label{fig:hmms}
\end{figure}

\section{Training Regime and Computational Resources}
\label{sec:training}
We conducted a systematic sweep across key architectural and task parameters to evaluate how RNNs replicate the emission statistics of different HMM families. Specifically, we trained networks on four \textit{linear-chain} HMMs ($M = 2, 3, 4, 5$ latent states) as well as on the \textit{fully-connected} and \textit{cyclic} architectures described in Section~\ref{subsec:data‑gen}. For the \textit{linear-chain} HMMs, we train networks with three different hidden state sizes ($|h| = 50, 150, 200$) and four input noise dimensionalities ($d = 1, 10, 100, 200$), yielding 48 unique configurations. Each configuration is trained with three independent random seeds, resulting in a total of 144 models. For the \textit{fully-connected} and \textit{cyclic} architectures, we fixed the hidden size to $|h| = 150$ and input dimensionality to $d = 100$, training three independent seeds per architecture, resulting in six additional models. Only seeds that successfully converged were retained for analysis.

The training regime is kept consistent across all configurations. Each RNN is trained on 30{,}000 sequences of fixed length, sampled from its corresponding HMM: 100 for $M=2$, 30 for $M=3,4$, 40 for $M=5$, 30 for the \textit{fully-connected}, and 40 for the \textit{cyclic} architectures. Optimization is performed in batches of 4096 using the Adam optimizer~\cite{kingma2014adam} with a learning rate of 0.001. Hidden states are initialized to zero at the start of training, and all weights are drawn from a uniform distribution \(\mathcal{U}(-\sqrt{k}, \sqrt{k})\), where \(k = \frac{1}{\texttt{hidden\_size}}\). To stabilize learning and mitigate exploding gradients, we apply gradient clipping with a maximum norm of 0.9 for the \textit{linear-chain} models and 0.3 for the \textit{fully-connected} and \textit{cyclic} architectures. Training proceeded until convergence, typically reached within 1000 epochs. However, convergence becomes less consistent as the number of hidden states in the target HMM increases, in which case shorter training sequences were used to facilitate convergence.

The complete pipeline—including HMM data generation, RNN training, computation of evaluation metrics, PCA analyses, and visualization—was optimized for efficient execution on modern hardware. On an NVIDIA RTX 4090 GPU, each model completed training in approximately 5–20 minutes, depending on sequence length and network size.

\section{Performance Metrics}
\label{sec:metrics}

To evaluate how well task-optimized RNNs replicate the behavior of the reference HMMs, we developed four performance metrics that capture both global sequence similarity and fine-grained statistical properties. These include: Euclidean distances between matched sequences, squared differences in transition matrices, observation frequencies, and observation volatility. Below we detail the definition, motivation, and implementation of each metric.

\paragraph{\textit{Sinkhorn-aligned} euclidean distance.}
To quantify the global discrepancy between HMM and RNN sequences, we compute the Euclidean distance after aligning predicted and reference sequences via the Sinkhorn divergence. This procedure ensures a principled pairing between sequences by solving a soft optimal transport problem, where matched sequences minimize expected pairwise distances under an entropy-regularized transport plan.

Given two sequences of categorical outputs, $\bm{y}_{\text{rnn}}$ and $\bm{y}_{\text{hmm}}$, represented as one-hot vectors in $\mathbb{R}^{T \times C}$ (where $T$ is sequence length and $C$ the number of output categories), we flatten them into vectors of dimension $T \cdot C$ and compute the Euclidean distance:

\[
    d(\bm{y}_{\text{rnn}}, \bm{y}_{\text{hmm}}) = \sqrt{\sum_{i=1}^{T \cdot C} \left( y_{\text{rnn}}^{(i)} - y_{\text{hmm}}^{(i)} \right)^2}
\]

\begin{figure}[h]
    \centering
    \includegraphics[width=1.0\textwidth]{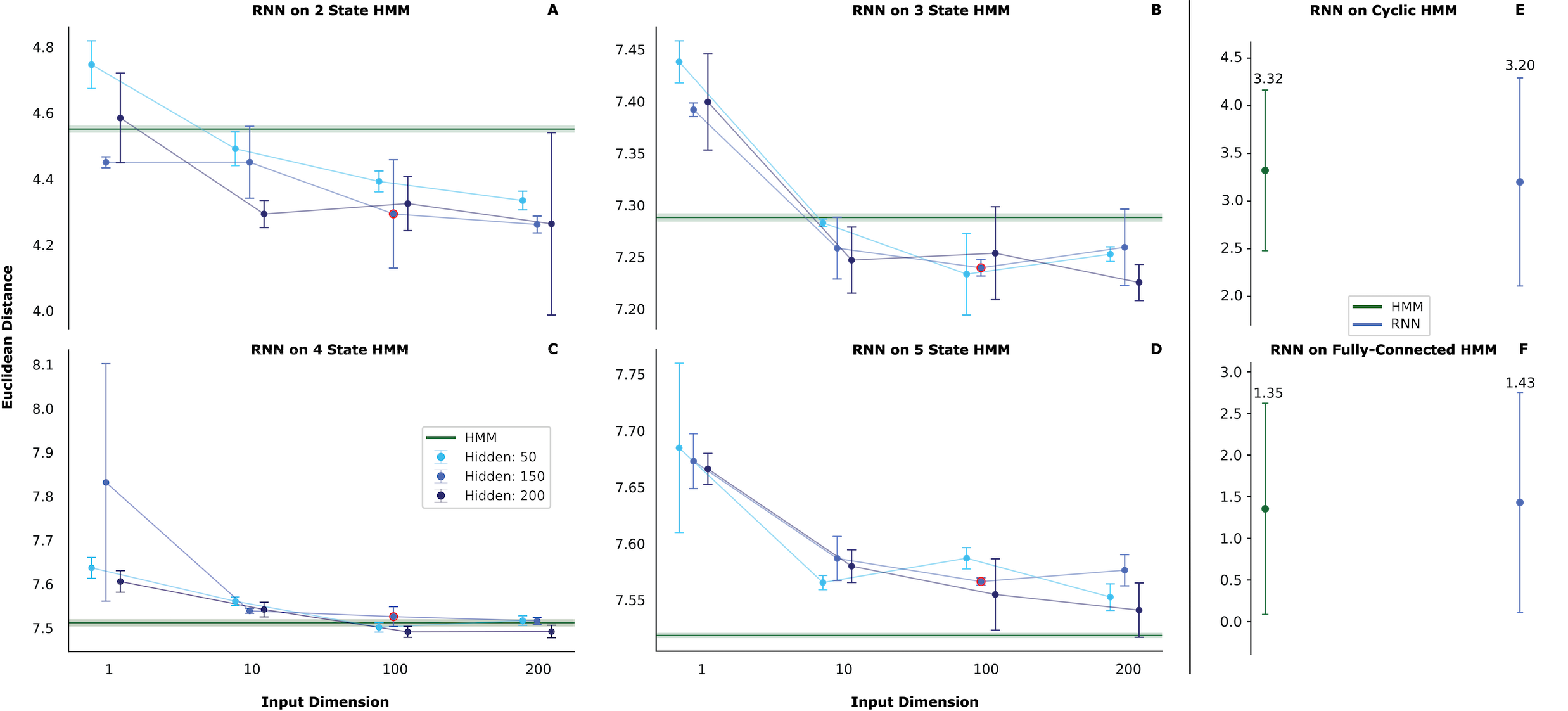}
    \caption{\textbf{Sinkhorn-aligned euclidean distances.} Mean $\pm$ s.d. Euclidean distances between Sinkhorn-aligned output sequences generated by trained RNNs and their corresponding HMM references, across all combinations of hidden-state size ($|h| \in {50,150,200}$) and input-noise dimensionality ($d \in {1,10,100,200}$). Panels \textbf{A–D} show results for \textit{linear-chain} HMMs with $M \in {2,3,4,5}$ latent states, while panels \textbf{E–F} correspond to the \textit{cyclic} and \textit{fully-connected} architectures, respectively. Error bars indicate variability across three random seeds per configuration. Horizontal green bands represent the baseline distance obtained by comparing HMM-generated sequences against themselves. Red-circled points mark the selected RNN configuration ($|h|=150$, $d=100$) used in the mechanistic analyses presented throughout the paper. For \textit{linear-chain} HMMs, increasing input dimensionality systematically improves alignment with the reference emissions, approaching the HMM baseline across hidden-state sizes. For \textit{fully-connected} and \textit{cyclic} architectures, RNNs trained with the same configuration ($|h|=150$, $d=100$) achieve similarly close correspondence to their HMM targets.}
    \label{fig:euc}
\end{figure}

Distances are computed across batches of $N = 5000$ sequences, averaged over random seeds and configurations. As a baseline, the same metric is computed between pairs of HMM-generated sequences. These distributions are visualized in Fig.~\ref{fig:euc}, demonstrating that RNN outputs closely match the HMM baseline.

\paragraph{Transition matrix squared differences.}
To assess how well the RNN captures the temporal dependencies between successive outputs, we compute empirical transition matrices from the sequences. For each model, we extract the most likely output at each time step and count the empirical frequency of transitions from output $i$ to $j$:

\[
T_{ij} = \frac{\# \text{ transitions from output } i \text{ to } j}{\# \text{ total transitions from } i}
\]

We compute $T^{\text{rnn}}$ and $T^{\text{hmm}}$, and compare them via element-wise squared differences:

\[
\Delta_{ij} = \left( T^{\text{rnn}}_{ij} - T^{\text{hmm}}_{ij} \right)^2
\]

\begin{figure}[h]
    \centering
    \includegraphics[width=1.0\textwidth]{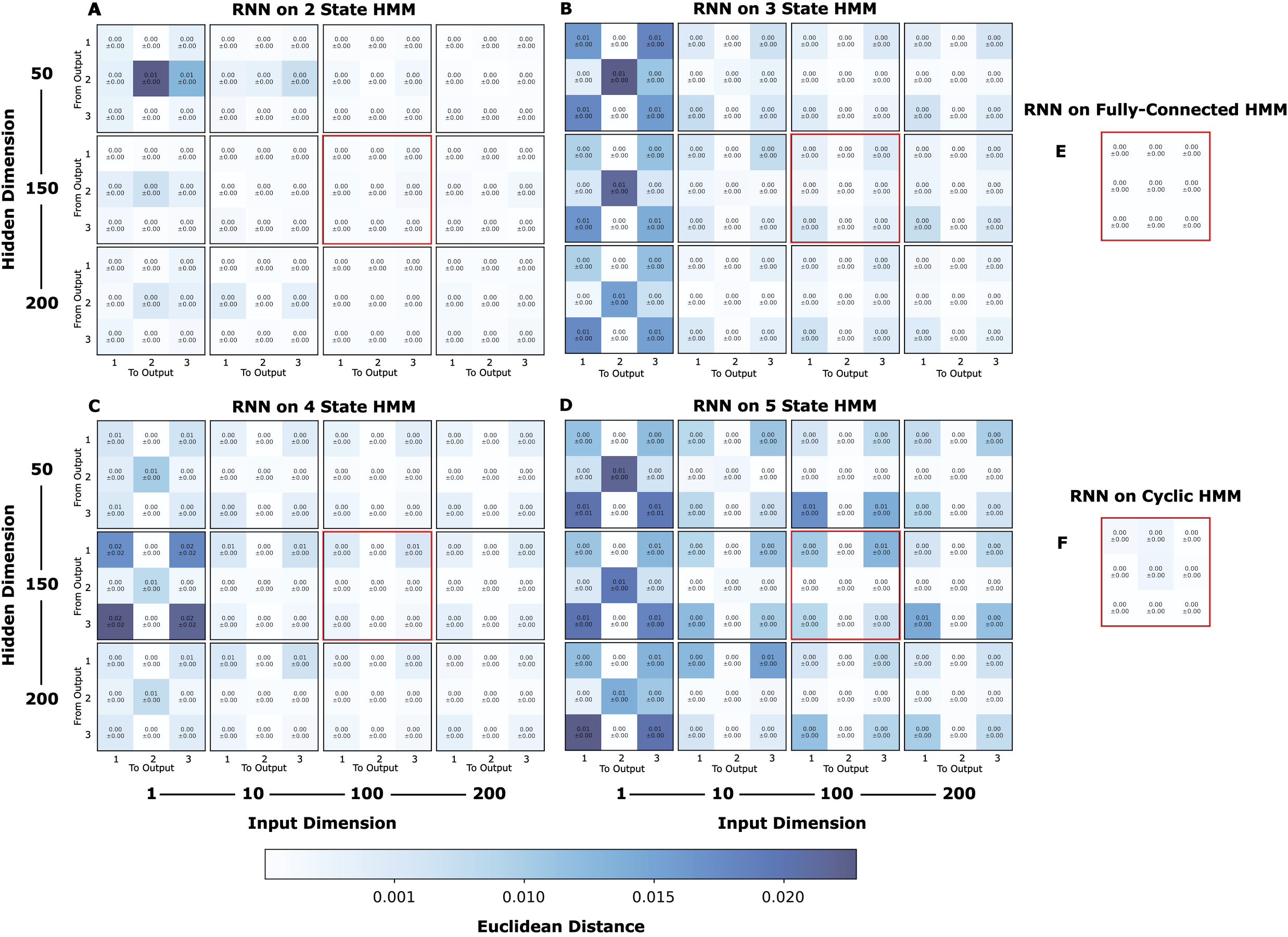}
    \caption{\textbf{Transition matrices squared differences.}Mean $\pm$ s.d. of the element-wise squared differences between empirical transition matrices computed from RNN outputs and those of their reference HMMs, across all combinations of hidden-state size ($|h| \in {50,150,200}$) and input-noise dimensionality ($d \in {1,10,100,200}$). Panels \textbf{A–D} correspond to \textit{linear-chain} HMMs with $M \in {2,3,4,5}$ latent states, while panels \textbf{E–F} report results for the \textit{fully-connected} and \textit{cyclic} architectures, respectively. Errors are averaged over three random seeds per configuration. Darker colors indicate larger deviations from the reference transition matrices, while lighter tones reflect closer alignment. Red-outlined boxes highlight the selected RNN configurations ($|h|=150$, $d=100$) used in the mechanistic analyses throughout the paper. For the \textit{linear-chain} architectures, transition-matrix alignment improves systematically with increasing input dimensionality, approaching near-perfect correspondence at $d \ge 100$. The \textit{fully-connected} and \textit{cyclic} models, trained with the same configuration, achieve comparably low deviation levels.}
    \label{fig:transition}
\end{figure}

The resulting matrices are averaged across models and shown in Fig.~\ref{fig:transition}, providing a detailed account of how accurately each RNN reproduces the internal transition structure of its target HMM.

\paragraph{Observation frequencies.}
This metric assesses the RNN's ability to reproduce the stationary distribution of HMM outputs. We count the frequency of each output class across all time steps and sequences, yielding a probability vector $p^{\text{rnn}} \in \mathbb{R}^C$. We compare this to the ground-truth distribution $p^{\text{hmm}}$ and, although not shown in figures, this analysis confirms that most trained networks match long-term HMM statistics closely.

\paragraph{Observation Volatility.}
To quantify short-term dynamics, we measure how frequently the RNN changes its output across time. Volatility is then averaged across sequences and compared with that of the HMM. This metric complements observation frequency by detecting under- or over-smoothing in the RNN's emission process.

\clearpage
\subsection{Sinkhorn-Aligned Euclidean Distances}

\clearpage
\subsection{Transition Matrices Squared Differences}

\clearpage
\section{Trajectories for RNNs trained on HMMs}
\label{sec:trajectories}
\subsection{Trajectories for Linear-Chain HMMs}
\begin{figure}[h]
    \centering
    \includegraphics[width=1.0\textwidth]{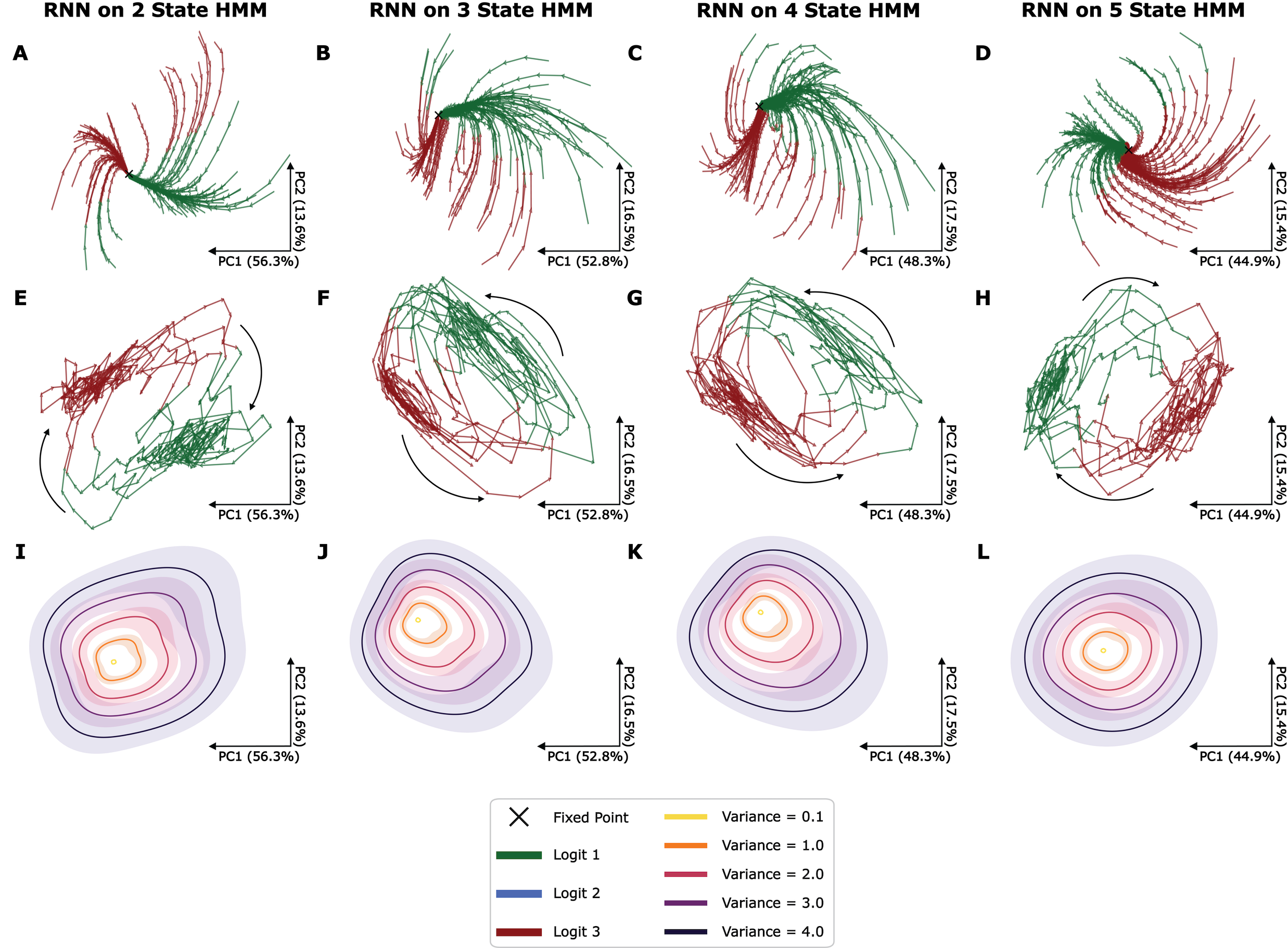}
    \caption{\textbf{Trajectories for linear-chain HMMs.} Each column shows results from RNNs ($|h| = 150$)  trained to reproduce the emission statistics of \textit{linear-chain} HMMs with \(M \in \{2, 3, 4, 5\}\) latent states. Hidden-state trajectories are projected onto the first two principal components (same axes across rows). \textbf{A–D}: trajectories from random initial conditions converge to a single fixed point (cross) in the absence of input. \textbf{E–H}: under Gaussian input noise, trajectories evolve along stable orbits with distinct regions corresponding to different dominant logits (colors). Arrows indicate average flow direction. \textbf{I–L}: hidden-state density contours (95\% CI) under increasing input variance (\(\sigma^2 \in \{0.1, 1.0, 2.0, 3.0, 4.0\}\)) reveal a linear scaling between orbit radius and input variance, while preserving overall shape.}
    \label{fig:trajectories_linearchain}
\end{figure}

\clearpage
\subsection{Trajectories for Fully-Connected and Cyclic HMMs}
\begin{figure}[h]
    \centering
    \includegraphics[width=1.0\textwidth]{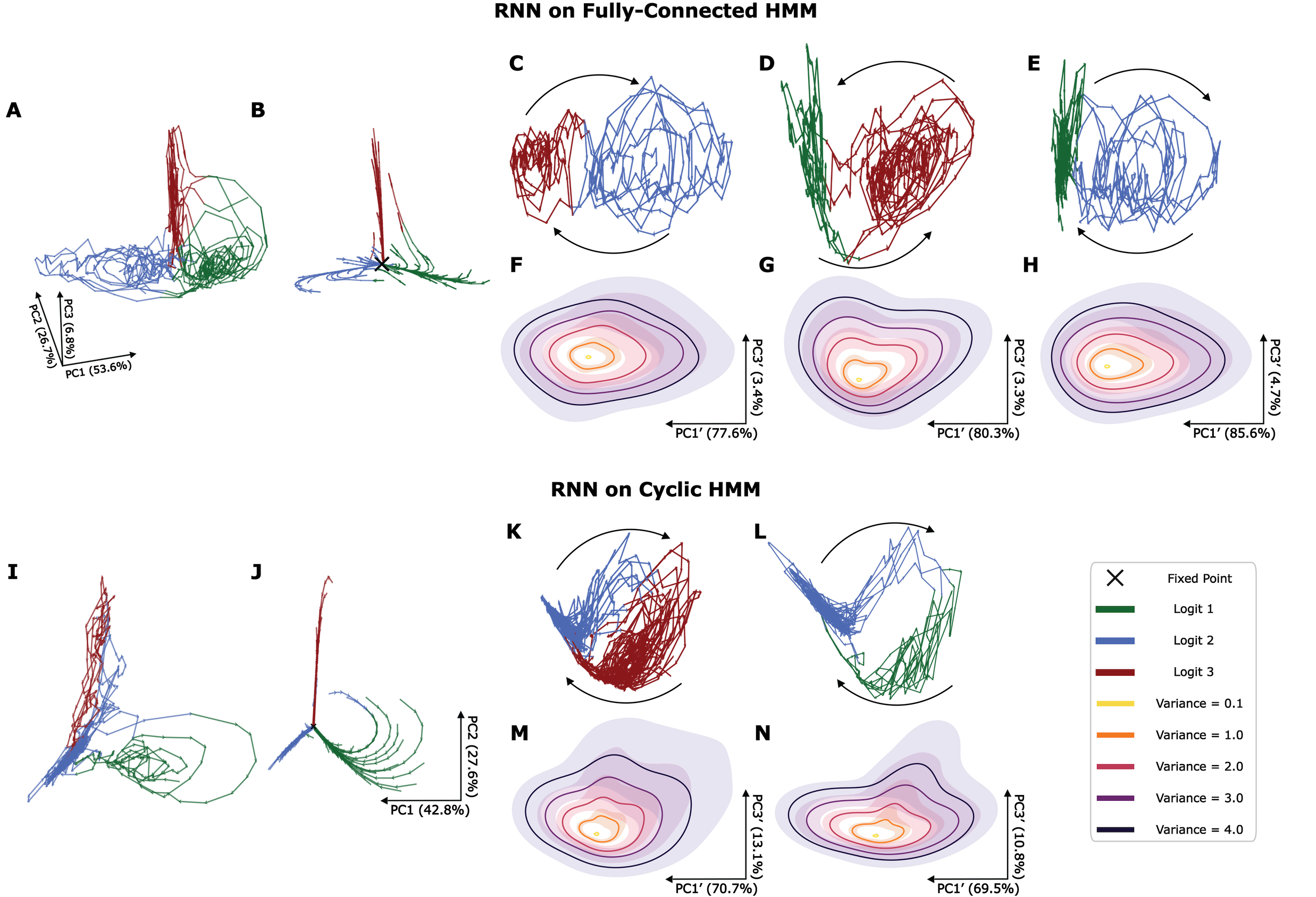}
    \caption{\textbf{Trajectories for fully-connected and cyclic HMMs.} Hidden-state dynamics of RNNs ($|h|=150$) trained to reproduce the emission statistics of \textit{fully-connected} (top) and \textit{cyclic} (bottom) HMMs.  Hidden-state trajectories are projected onto the first three principal components in \textbf{A–B} and \textbf{I–J}. To visualize orbital dynamics linking any given pair of clusters, within the subspace in \textbf{A–B} and \textbf{I–J} we computed a second set of PC component separately for activity restricted to each pair, shown in \textbf{C–H} and \textbf{K–N}. \textbf{B, J}: Under Gaussian input noise, trajectories evolve along multiple stable orbits connecting slow regions dominated by distinct logits (colors). These orbital trajectories can be decomposed into the same fundamental dynamical primitives described in the text, shown in \textbf{C–E} and \textbf{K–L}, respectively. \textbf{A, I}: In the absence of input, trajectories from random initial conditions converge to a single fixed point (cross). \textbf{F–H} and \textbf{M–N}: Hidden-state density contours (95 \% CI) under increasing input variance (\(\sigma^2 \in \{0.1, 1.0, 2.0, 3.0, 4.0\}\)) reveal that the linear scaling between input variance and orbit radius is preserved.}
    \label{fig:trajectories_fullyconnected}
\end{figure}

\clearpage
\section{Learning Trajectories Across Training Epochs}
\label{sec:learning_traj}
\begin{figure}[h]
    \centering
    \includegraphics[width=1.0\textwidth]{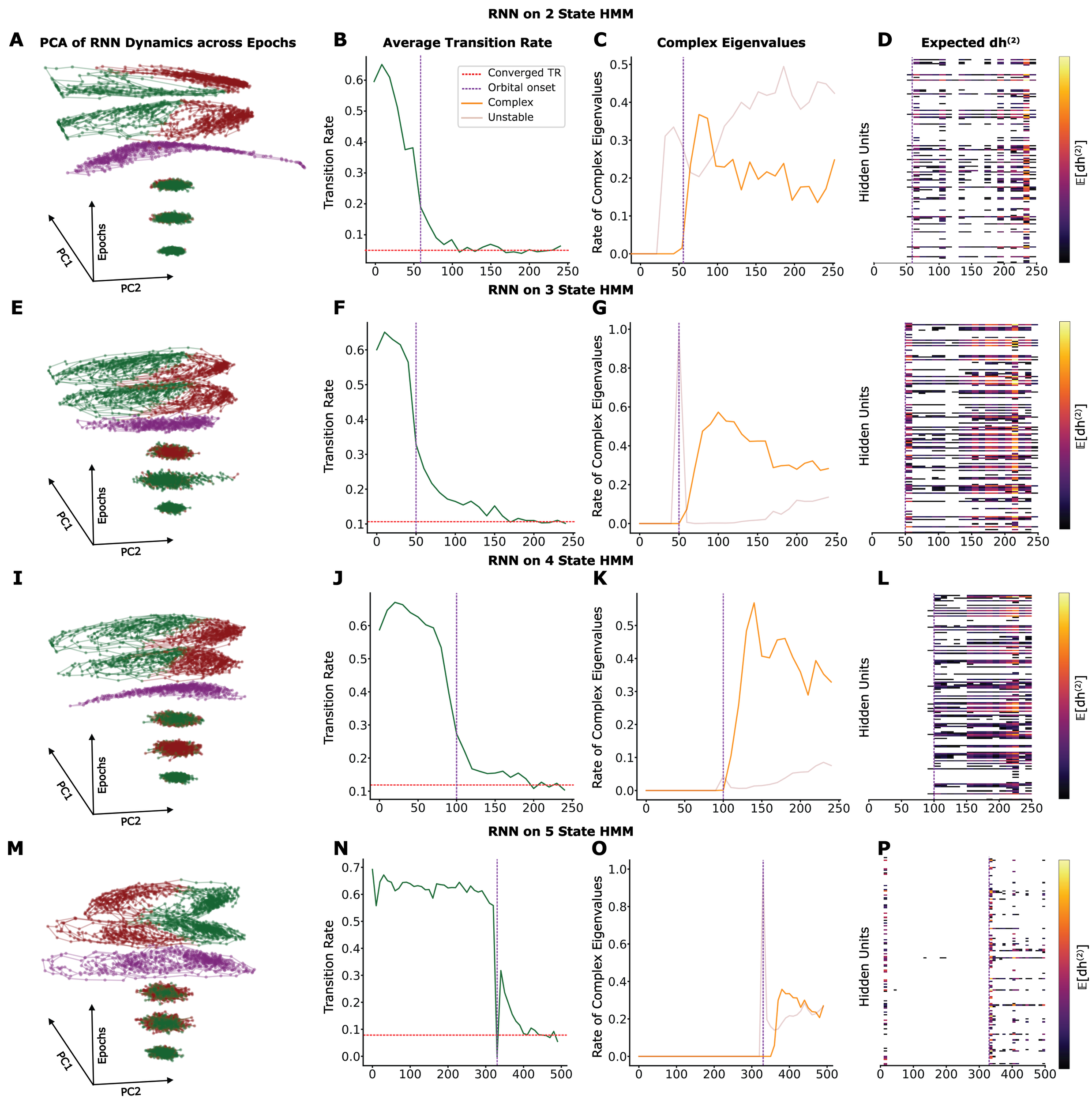}
    \caption{\textbf{Learning trajectories across training epochs for linear-chain HMMs.} Each row corresponds to an RNN trained on a \textit{linear-chain} HMM with \(M \in \{2, 3, 4, 5\}\) latent states (top to bottom). \textbf{A, E, I, M}: PCA projections of hidden-state dynamics across training epochs reveal the emergence of structured orbital dynamics: networks initially converge to a single fixed point, undergo a transient unstable regime (purple), and ultimately form stable, noise-sustained orbital dynamics (colored by dominant logit). \textbf{B, F, J, N}: the average transition rate between output clusters, showing a clear shift after the transition epoch (purple dashed line) aligning with the converged transition rate (red dotted line). \textbf{C, G, K, O}: the fraction of unstable and complex Möbius-transformed eigenvalues of the Jacobian rises sharply around the transition, reflecting, respectively, the destabilization of the fixed point and the onset of oscillatory dynamics. \textbf{D, H, L, P}: the expected second-order perturbation vector \(\mathbb{E}[dh^{(2)}]\) emerges after this transition, capturing how input noise variance drives the recurrent dynamics and sustains the orbital regime. Together, these results demonstrate a consistent learning trajectory across architectures, in which the network transitions from stable to oscillatory dynamics.}
    \label{fig:learning_traj}
\end{figure}

\clearpage
\begin{figure}[h]
    \centering
    \includegraphics[width=1.0\textwidth]{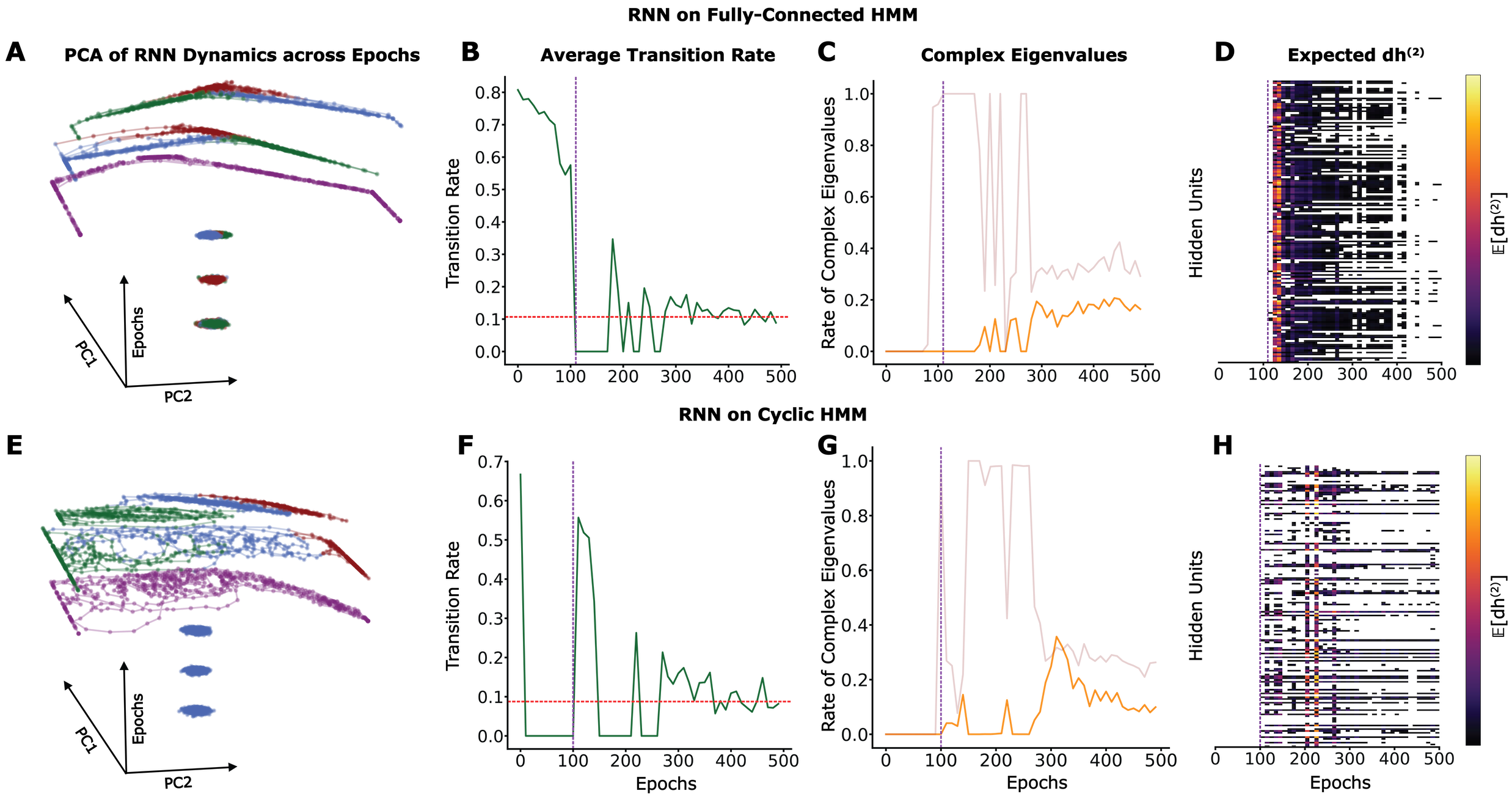}
    \caption{\textbf{Learning trajectories across training epochs for fully-connected and cyclic HMMs.} Each row corresponds to an RNN ($|h| = 150$) trained to reproduce the emission statistics of a \textit{fully-connected} (top) or \textit{cyclic} (bottom) HMM. \textbf{A, E}: PCA projections of hidden-state dynamics across training epochs show the progressive emergence of structured orbital dynamics: networks initially converge to a single fixed point, undergo a transient unstable regime (purple), and ultimately form stable, noise-sustained orbital dynamics (colored by dominant logit). \textbf{B, F}: the average transition rate between output clusters, showing a clear shift after the transition epoch (purple dashed line) aligning with the converged transition rate (red dotted line).  \textbf{C, G}: the fraction of unstable and complex Möbius-transformed eigenvalues of the Jacobian rises sharply around the transition, reflecting, respectively, the destabilization of the fixed point and the onset of oscillatory dynamics. \textbf{D, H}: the expected second-order perturbation vector \(\mathbb{E}[dh^{(2)}]\) emerges after this transition, capturing how input noise variance drives the recurrent dynamics and sustains the orbital regime. These results demonstrate that RNNs trained on more complex HMMs follow the same qualitative learning trajectory observed for linear-chain models, transitioning from a single attractor to oscillatory dynamics.}
    \label{fig:learning_traj2}
\end{figure}

\clearpage
\section{Loss Curves}
\begin{figure}[h]
    \centering
    \includegraphics[width=1.0\textwidth]{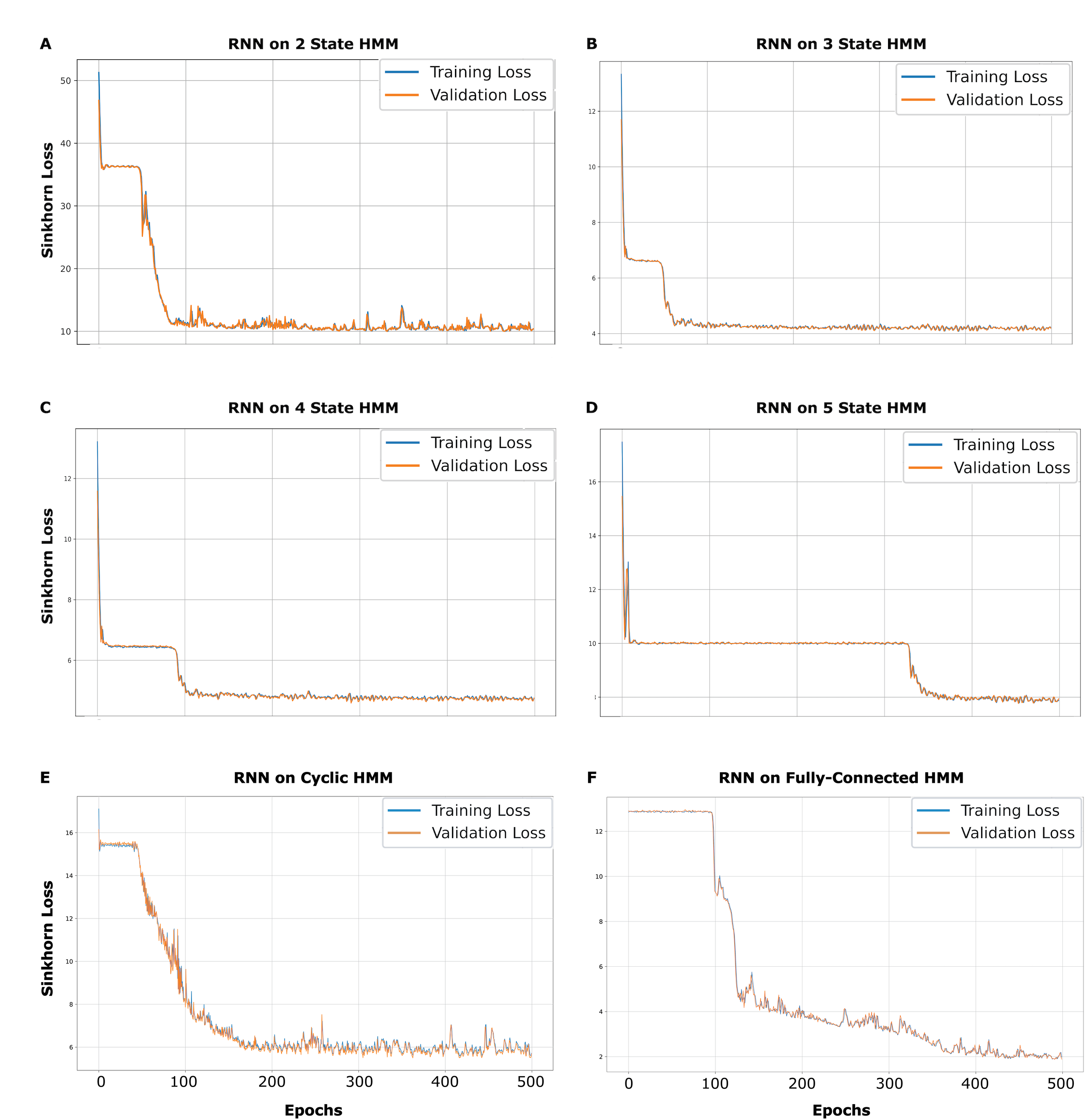}
    \caption{\textbf{Training and validation loss curves.} Sinkhorn loss over training epochs for the six RNN models used in the main analyses (\(|h| = 150\), \(d = 100\)), each trained to reproduce the emission statistics of an \textit{HMM} with different transition structures: \textit{linear-chain} \(M \in \{2, 3, 4, 5\}\), \textit{fully-connected}, and \textit{cyclic}. Training and validation losses (blue and orange) closely overlap across all configurations, indicating no signs of overfitting. All models converge within 500 epochs, exhibiting a characteristic \textit{double-descent} profile. The second loss drop coincides with the transient unstable regime identified in Figures~\ref{fig:learning_traj} - \ref{fig:learning_traj2}, marking the emergence of noise-sustained orbital dynamics and the transition from fixed-point to rotational behavior. This correspondence confirms that the onset of orbital dynamics constitutes the dynamical solution through which the networks minimize the Sinkhorn loss.}
    \label{fig:loss}
\end{figure}

\clearpage
\section{Jacobian Linearization and Stability Analysis via Möbius Transformation}
\label{app:jacobian}
In a vanilla RNN with ReLU non‑linearity the hidden state evolves as
\begin{equation}
    h_t \;=\; \phi\!\bigl(x_t\,W_{\text{in}}^T + h_{t-1}\,W_{hh}^T\bigr),
    \qquad \phi(z)=\max(0,z).
    \tag{A.1}
\end{equation}
Because the derivative of the ReLU is either $0$ or $1$, the Jacobian that
propagates an infinitesimal perturbation $\delta h_{t-1}$ to $\delta h_t$ takes
a particularly simple form:
\begin{equation}
    J_t \;=\; W_{hh}^TD_t ,
    \qquad
    D_t \;=\; \operatorname{diag}\!\bigl[\mathbbm 1_{z_t>0}\bigr],
    \qquad
    z_t = x_t\,W_{\text{in}}^T + h_{t-1}\,W_{hh}^T.
    \tag{A.2}
\end{equation}
The diagonal \emph{gating matrix} $D_t$ switches rows of $W_{hh}$ on or off
depending on whether the corresponding pre‑activation is positive.

\paragraph{Discrete‑time stability and the need for a spectral map.}
In discrete time a fixed point is locally stable iff all eigenvalues satisfy
$|\lambda_i|<1$.  To interpret this spectrum with the intuition of continuous‑time
systems (where the relevant boundary is $\operatorname{Re}\lambda_i=0$) we apply
the classical Möbius transformation
\begin{equation}
    \mu(\lambda) \;=\; \frac{1+\lambda}{1-\lambda},
    \qquad \lambda \neq 1.
    \tag{A.4}
\end{equation}
This bijection maps the unit circle to the imaginary axis and preserves
complex conjugacy, giving the correspondence
\begin{center}
\begin{tabular}{@{}lll@{}}
\toprule
Condition in $\lambda$ & After mapping & Interpretation \\\midrule
$|\lambda|<1$  & $\operatorname{Re}\mu(\lambda)<0$ & contraction \\
$|\lambda|=1$  & $\operatorname{Re}\mu(\lambda)=0$ & neutral / Hopf boundary \\
$|\lambda|>1$  & $\operatorname{Re}\mu(\lambda)>0$ & expansion \\
\bottomrule
\end{tabular}
\end{center}

\clearpage
\section{Second Order Perturbation}
\label{sec:second order perturbation}
Inspired by the perturbative approach on RNN of \cite{lim2021understanding}, here we compute two simple terms to estimate the impact of the first and second order effects of the input noise on the stable dynamics. Without loss of generality, we reformulate the RNN dynamics as:

$$\hat{h}_{t+1} = \phi(\hat{h}_t)W_{hh}^T + x_tW_{\text{in}}^T\epsilon$$

With $\phi$ representing the ReLU activation, $\hat{h}_t \in \mathbb{R}^r$ the pre-activation recurrent state, $x_t \in \mathbb{R}^n$ with $x_t \sim \mathcal{N}(0,\sigma^2\mathbb{I}_n)$, and $z_i$ the $i-th$ pre-activation neuron. We recover the formulation of equation (2) in Section 3.2 with $h_t = \phi(\hat{h}_t)$. Let's define the unperturbed dynamics
$$\hat{h}^o_t = \phi(\hat{h}^o_{t-1})W_{hh}^T$$
Such that $\hat{h}_{t_c}^0 = \hat{h}_{{t_c}-1}^0$ for some $t_c >0$. We consider the first order perturbation around it
$$
\hat{h}_t = \hat{h}_t^0 + \delta h_t
$$
By substituting we obtain the following:

$$
\hat{h}_{t+1}^0 + \delta h_{t+1} = \phi(\hat{h}_t'^0 + \delta h_t)W_{hh}^T +  x_t W_{\text{in}}^T\epsilon,
$$

$$
\delta h_{t+1} = \phi(\hat{h}_t^0 + \delta h_t)W_{hh}^T - \phi(\hat{h}_t^0)W_{hh}^T + x_t W_{\text{in}}^T \epsilon.
$$

Given the ReLU activation, we approximate for small $\delta h_t$:

$$
\phi(\hat{h}_t^0 + \delta h_t) \approx \phi(\hat{h}_t^0) + \delta h_tD_\phi(\hat{h}_t^0),
$$

where $D_\phi(\hat{h}_t^0)$ is diagonal, with entries $1$ if $(\hat{h}_t^0)_i > 0$, $0$ otherwise. Thus:

$$
\delta h_{t+1} \approx  \delta h_t D_\phi(\hat{h}_t^0)W_{hh}^T + x_t W_{\text{in}}^T \epsilon.
$$

Naming $A_t = D_\phi(\hat{h}_t^o)W_{hh}^T$ and $M_{t,s} = A_{s+1} A_{s+1} ...A_t = \prod_{k=s+1}^{t+1} A_k$, with $A_{t+1}=\mathbb{I}_r$. 

By iterating we obtain:
$$
\delta h_t = \epsilon \sum_{s=0}^{t} x_s W_{\text{in}}^T \left( \prod_{k=s+1}^{t+1}A_k \right)  =  \epsilon\sum_{s=0}^{t} x_s W_{\text{in}}^TM_{t,s} 
$$
$$
\hat{h}_t = \hat{h}_t^o + \epsilon \sum_{s=0}^{t} x_s W_{\text{in}}^T M_{t,s} 
$$
We observe that, given the unbiased Gaussian noise, $\mathbb{E}[\delta h] = 0$. Therefore, the RNN integration mechanism must rely on higher order terms. Henceforth, we herby derive the second order term as well. To make the higher-order term explicit, Instead of $\delta h = \delta h^{(1)} + O(\sigma^2)$ we express $\delta h_t$ into orders of noise:

$$
\delta h_t = \delta h_t^{(1)} + \delta h_t^{(2)} + O(\sigma^3),
$$
where $\delta h_t^{(1)}$ is the first-order term and $\delta h_t^{(2)}$ is the second-order term. For an element-wise activation ($\frac{\partial^2 \phi_i}{\partial z_j^2} = 0$ for $i \neq j$), we get the following recurrent form

$$\delta h_{t+1} =  \delta h_t D_\phi( \hat{h}_t^0) W_{hh}^T + \frac{1}{2} \sum_{i=1}^r \left[ (\delta h_t)_i^2 \cdot \frac{\partial^2 \phi_i}{\partial z_i^2}(\hat{h}_t^0) \right] e_i \cdot W_{hh}^T + x_t W_{\text{in}}^T  + O(\|\delta h_t\|^3) $$

Given that $\delta h_t^{(2)}$ captures only the second order effects, it must satisfy the following relation

$$
\delta h_{t+1}^{(2)} \approx \delta h_t^{(2)} D_\phi( \hat{h}_t^0) W_{hh}^T+ \frac{1}{2} \sum_{i=1}^r \left[ (\delta h_t^{(1)})_i^2 \cdot \frac{\partial^2 \phi_i}{\partial z_i^2}(\hat{h}_t^0) \right] e_i \cdot W_{hh}^T
$$

with $\delta h_0^{(2)} = 0$. This is a linear recursion, driven also by a quadratic term in $\delta h_t^{(1)}$. Solving it we obtain:
$$
\delta h_t^{(2)} = \frac{\epsilon^2}{2}  \sum_{s=0}^{t}\sum_{i=1}^r \left[ (\delta h_s^{(1)})_i^2 \cdot \frac{\partial^2 \phi_i}{\partial z_i^2}(\hat{h}_s^0) \right] e_i \cdot M_{t,s} 
$$

Which leads to the following expectation, depending only on the noise variance.
$$
\mathbb{E}[\delta h_t^{(2)}] =  \frac{\epsilon^2}{2}\sum_{s=0}^{t} \sum_{i=1}^r \left[ \sigma^2\left(  \sum_{k=0}^{s} W_{\text{in}}^T M_{t,
s}    \right) _i^2 \cdot \frac{\partial^2 \phi_i}{\partial z_i^2}(\hat{h}_s^0) \right]e_i \cdot M_{t,s} 
$$

Unfortunately, the ReLU activation does not possess a second derivative. Henceforth, we approximate the impact of the second order perturbation with the square of the first order perturbation over several trials whenever the pre-activation neurons activity $z_i$ is below $0$:

$$dh^{(2)}_t = \sum^t_k \frac{1}{2} \left (dh^{(1)}_k \right)^2 \odot f_k \cdot M_{t,k}$$
Where $\odot$ is the Hadamard product and 

$$ (f_k)_i = \left\{\begin{matrix}
1 \text{ if }(z_k)_i<0
 \\
0 \text{ otherwise }\end{matrix}\right.$$

for t = 10 and averaging with respect to 100 different trajectories. This vector emerges after the orbital dynamics onset (Figure \ref{fig:cactus}A,D) and scales linearly with the variance, analogous to the scaling of the orbits observed in Figure \ref{fig:traj}D, K-M. Furthermore, empirically, we observe that kick neurons comprise some of the top non-zero components of this vector, suggesting a deeper connection between variance and the firing mechanism.

\clearpage
\section{Residency times and noise sensitivity}
\label{sec:residency_graph}
\begin{figure}[h]
    \centering
    \includegraphics[width=1.0\textwidth]{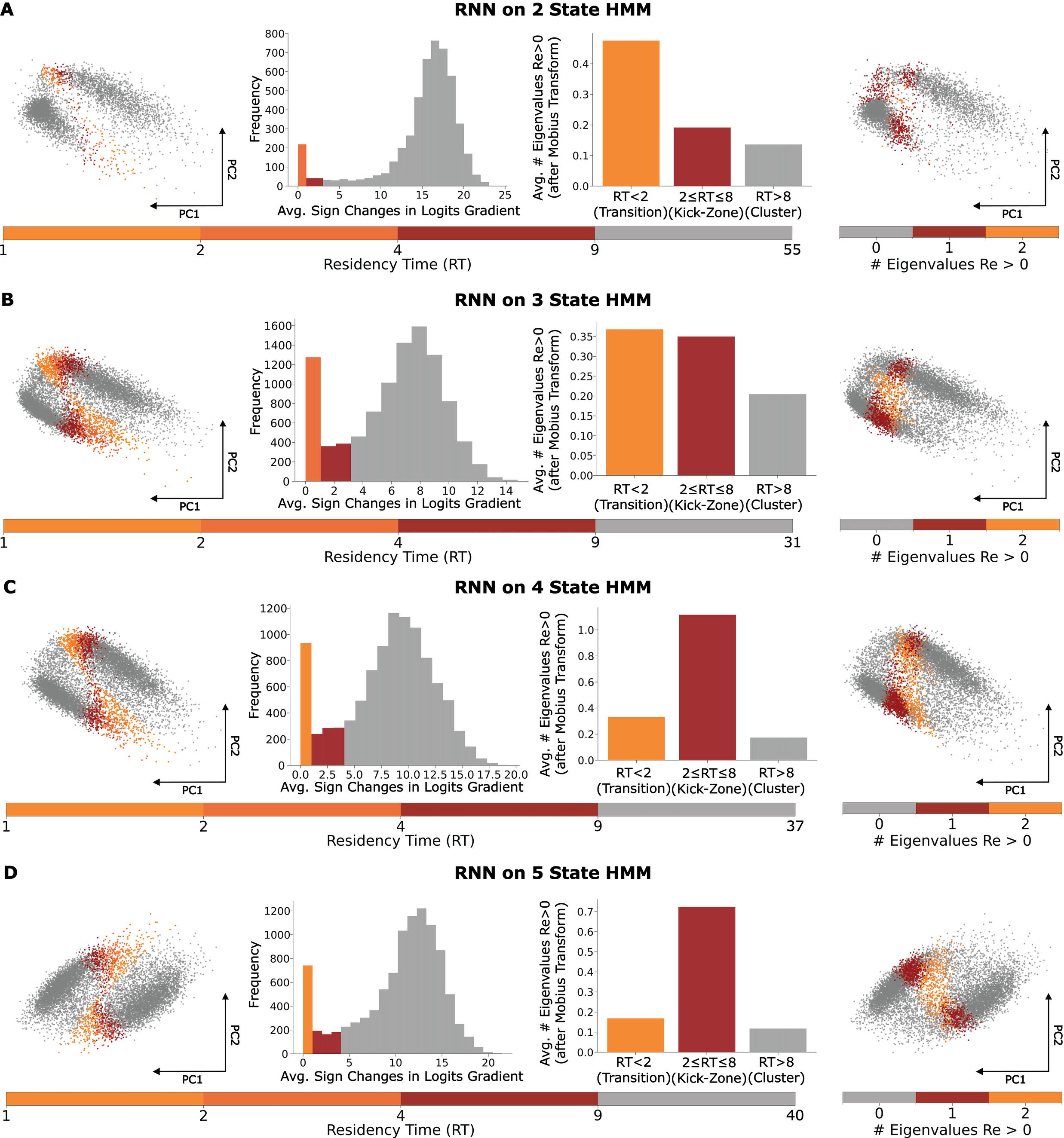}
    \caption{\textbf{Residency analyses for linear-chain HMMs.} Each row corresponds to an RNN trained on a \textit{linear-chain} HMM with \(M \in \{2, 3, 4, 5\}\) latent states (panels \textbf{A–D}). \textbf{Left}: PCA projections of the latent trajectories colored by residency time (RT) reveal distinct dynamical regimes, with slow regions (\textit{clusters}, dark gray) where trajectories linger and fast regions (\textit{transitions}, orange) where rapid output switching occurs. \textbf{Center-left}: distributions of average sign changes in the logit gradient exhibit a robust bimodal structure separating stable clusters from directed transition flows. \textbf{Center-right}: the average number of unstable directions (via Möbius-transformed Jacobian eigenvalues) peaks in intermediate RTs, identifying \textit{kick-zones} that mediate transitions. \textbf{Right}: spatial maps of the number of eigenvalues with positive real part confirm local instability confined to these zones. The coherent structure across all HMM configurations demonstrates that the RNNs converge on a common solution: orbital dynamics divided into slow clusters, unstable kick-zones, and rapid transitions, suggesting a generic mechanism by which RNNs can emulate discrete stochastic HMMs emissions through continuous dynamics.}
    \label{fig:residency_graph1}
\end{figure}

\clearpage

\begin{figure}[h]
    \centering
    \includegraphics[width=1.0\textwidth]{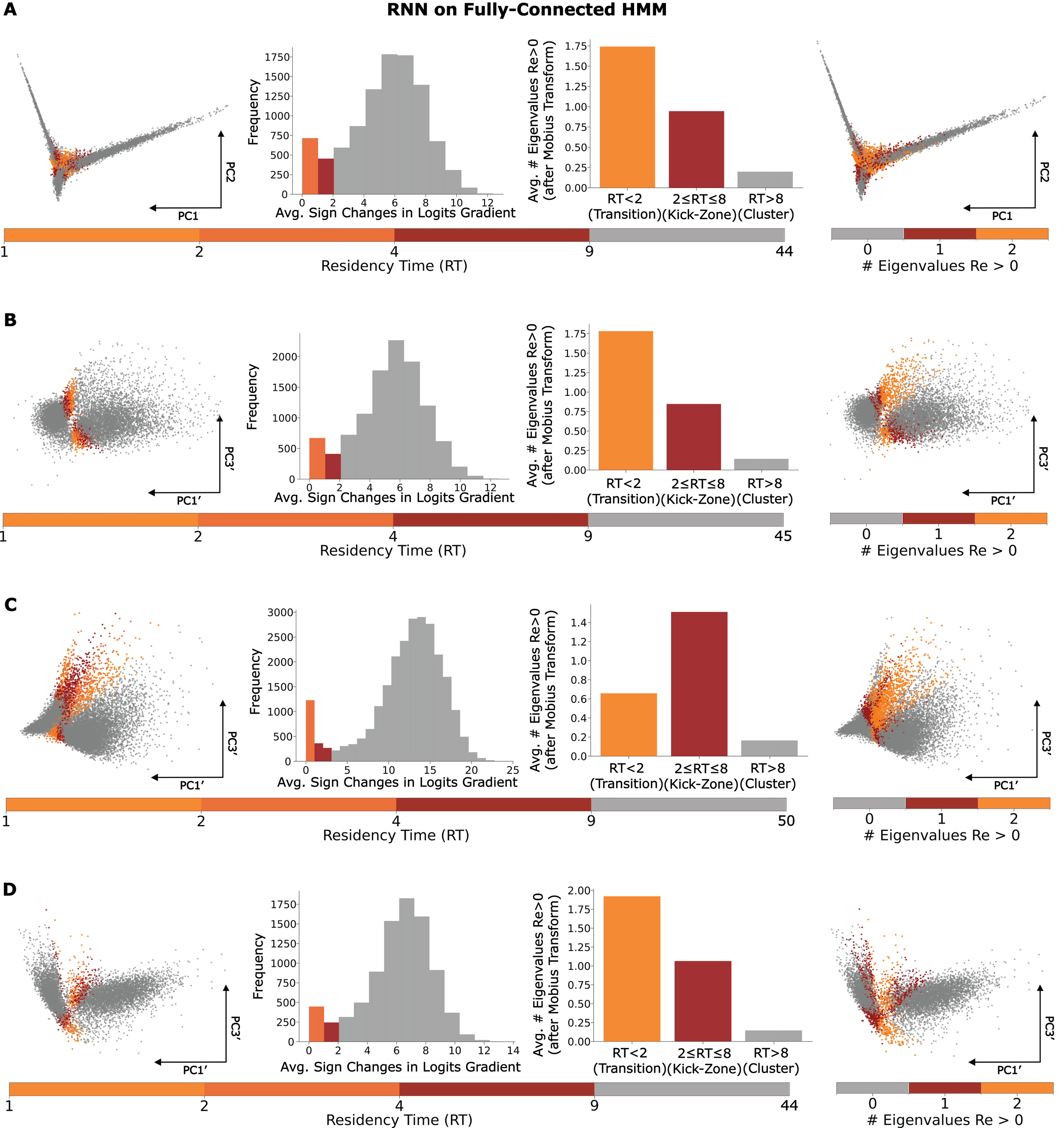}
    \caption{\textbf{Residency analyses for fully-connected HMMs.} Each row corresponds to the principal space or one of the three principal subspaces identified in the RNN trained to reproduce the emission statistics of a \textit{fully-connected} HMM (panels \textbf{A–D}). \textbf{Left}: PCA projections of the latent trajectories colored by residency time (RT) reveal distinct dynamical regimes, with slow regions (\textit{clusters}, dark gray) where trajectories linger and fast regions (\textit{transitions}, orange) where rapid output switching occurs. \textbf{Center-left}: distributions of average sign changes in the logit gradient exhibit a robust bimodal structure separating stable clusters from directed transition flows. \textbf{Center-right}: the average number of unstable directions, computed from Möbius-transformed Jacobian eigenvalues, peaks in intermediate RTs, identifying locally unstable \textit{kick-zones} that mediate transitions.  \textbf{Right}: spatial maps of the number of eigenvalues with positive real part confirm that instability is confined to kick-zones while clusters remain locally stable. Together, these results show that the RNN decomposes the fully-connected HMM into three coupled dynamical subspaces, each expressing the same tripartite organization of \textit{clusters}, \textit{kick-zones}, and \textit{transitions} described for the linear-chain architectures — supporting the compositional reuse of the same dynamical primitive across HMM families.}
    \label{fig:residency_graph2}
\end{figure}

\clearpage

\begin{figure}[h]
    \centering
    \includegraphics[width=1.0\textwidth]{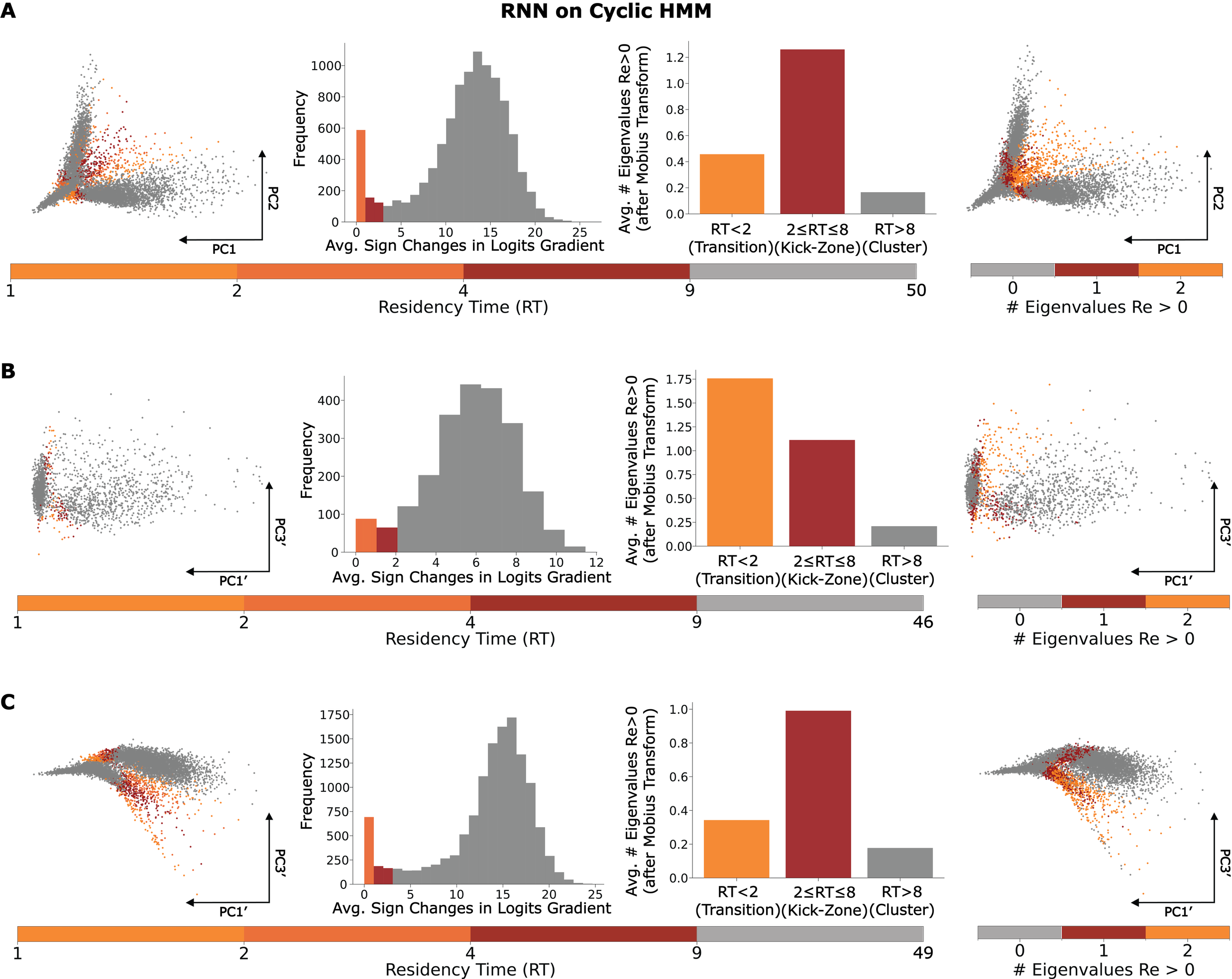}
    \caption{\textbf{Residency analyses.} Each row corresponds to the principal space or one of the two principal subspaces identified in the RNN trained to reproduce the emission statistics of a \textit{cyclic} HMM (panels \textbf{A–C}). \textbf{Left}: PCA projections of the latent trajectories colored by residency time (RT) reveal distinct dynamical regimes, with slow regions (\textit{clusters}, dark gray) where trajectories linger and fast regions (\textit{transitions}, orange) where rapid output switching occurs. \textbf{Center-left}: distributions of average sign changes in the logit gradient exhibit a robust bimodal structure separating stable clusters from directed transition flows. \textbf{Center-right}: the average number of unstable directions, computed from Möbius-transformed Jacobian eigenvalues, peaks in intermediate RTs, identifying locally unstable \textit{kick-zones} that mediate transitions.  \textbf{Right}: spatial maps of the number of eigenvalues with positive real part confirm that instability is confined to kick-zones while clusters remain locally stable. Together, these results show that cyclic architectures preserve the same dynamical primitive observed in simpler HMMs.}
    \label{fig:residency_graph3}
\end{figure}

\clearpage

\begin{figure}[h]
    \centering
    \includegraphics[width=1.0\textwidth]{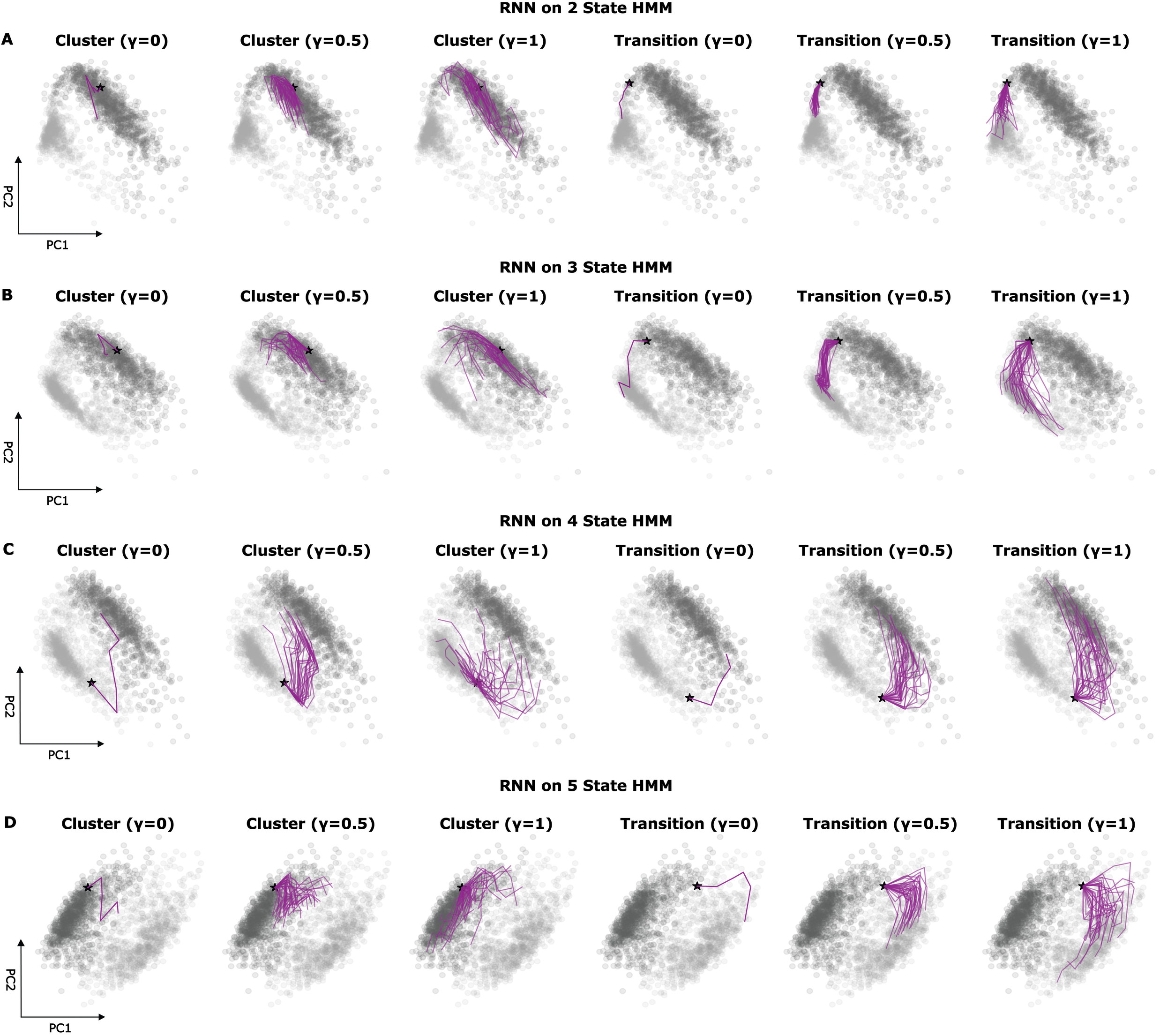}
    \caption{\textbf{Noise sensitivity analyses for linear-chain HMMs.} Each row corresponds to an RNN trained to reproduce the emission statistics of a \textit{linear-chain} HMM with \(M \in \{2, 3, 4, 5\}\) latent states (panels \textbf{A–D}). \textbf{Left to right:} Sampled initial conditions (black stars) are taken from representative \textit{cluster} (left triplets) and \textit{transition} (right triplets) regions in the latent space. From each location, 30 trajectories (magenta) are simulated under three levels of input-noise resampling: identical (\(\gamma = 0\)), partially resampled (\(\gamma = 0.5\)), and fully independent (\(\gamma = 1\)). In all configurations, transitions remain compact and exhibit robust, quasi-deterministic flow once the kick-zone is crossed, whereas clusters show increasing divergence with noise resampling, reflecting high noise sensitivity.}
    \label{fig:sensitivity}
\end{figure}

\clearpage

\begin{figure}[h]
    \centering
    \includegraphics[width=1.0\textwidth]{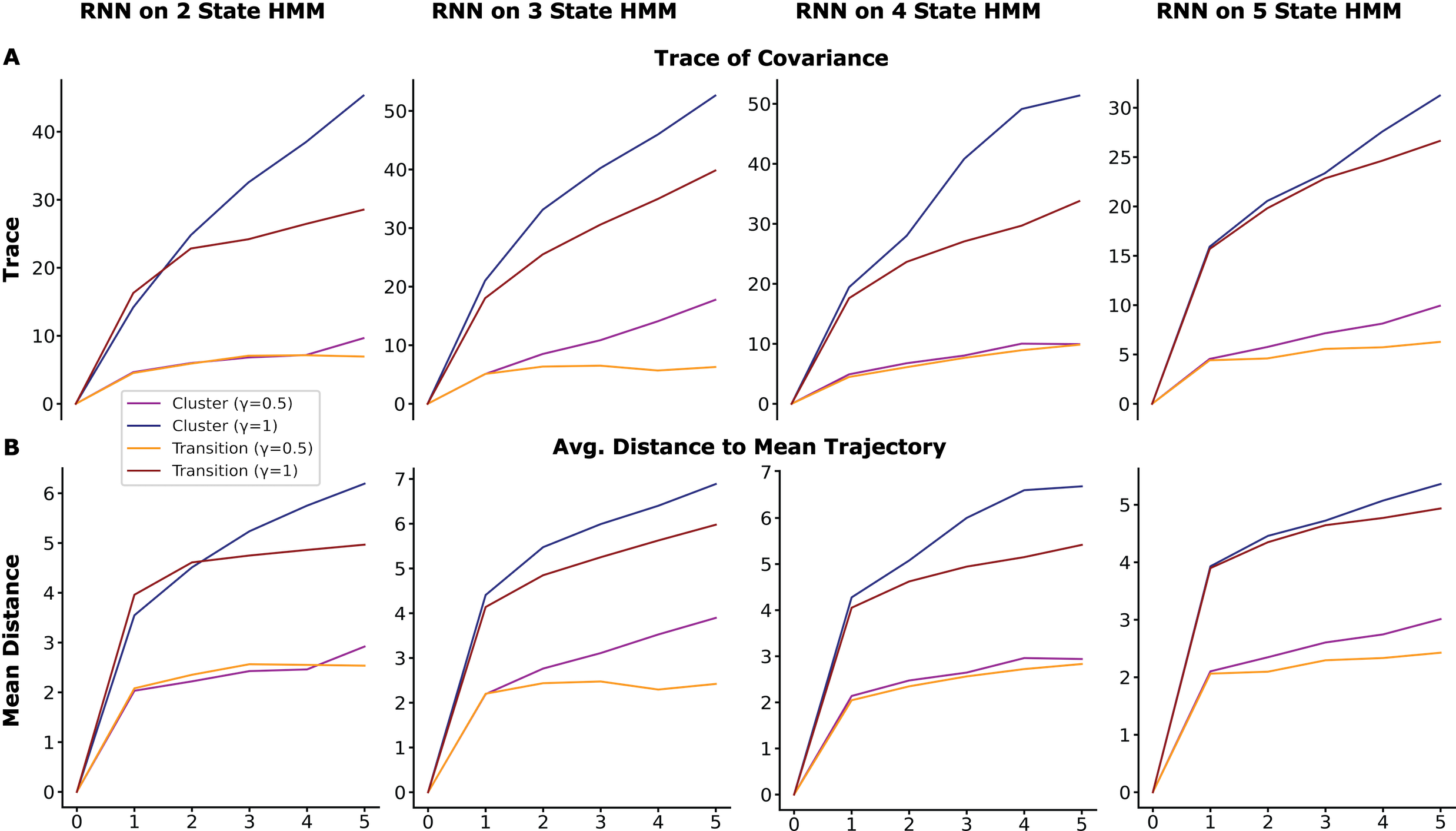}
    \caption{\textbf{Quantification of noise sensitivity for linear-chain HMMs.} Each column corresponds to an RNN trained to reproduce the emission statistics of a \textit{linear-chain} HMM with \(M \in \{2, 3, 4, 5\}\) latent states (panels \textbf{A–D}). For each configuration, we quantify the dispersion of latent trajectories initialized from representative \textit{cluster} (purple and blue) and \textit{transition} (orange and red) regions under two levels of noise resampling (\(\gamma = 0.5,\ 1\)). \textbf{A}: the \textit{trace of the trajectory covariance matrix} captures the overall spread of trajectories in latent space. \textbf{B}: the \textit{average Euclidean distance to the mean trajectory} reflects deviations across trials. In all configurations, transitions remain compact and exhibit robust, quasi-deterministic flow once the kick-zone is crossed, whereas clusters show progressively larger divergence with increasing noise resampling, reflecting high noise sensitivity.}
    \label{fig:trace}
\end{figure}

\clearpage

\section{Kick-Neurons Activity}
\label{sec:neurons}
\begin{figure}[h]
    \centering
    \includegraphics[width=1.0\textwidth]{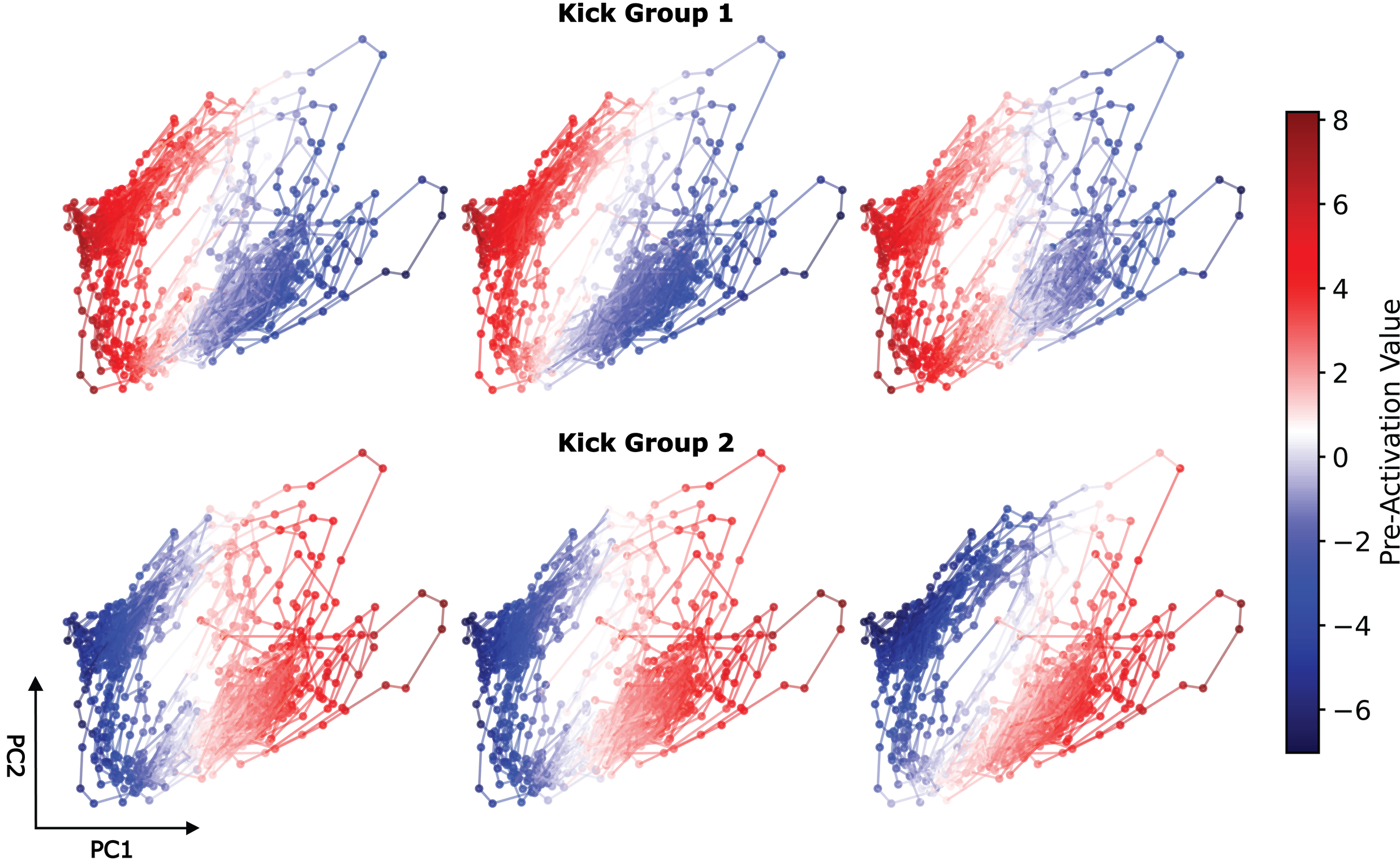}
    \caption{\textbf{Pre-activation values of kick-neurons.} Latent trajectories projected onto the first two principal components (PC1 and PC2) of the RNN trained on a 2-state HMM. Each point along a trajectory corresponds to one time step and is colored by the pre-activation value of a specific neuron. The first row shows the three neurons forming \textit{Kick Group 1}, while the second row displays the three neurons of \textit{Kick Group 2}, which mediate transitions in the opposite direction. In both groups, pre-activations are strongly negative within one of the two clusters, increase to near-zero values in the kick-zones, and become positive during transitions and in the subsequent cluster. These dynamics reveal a clear functional role: within clusters, the neurons are fully suppressed (ReLU outputs zero), while in the kick-zones, pre-activation values fluctuate near the ReLU threshold. This intermediate regime places the units near the \textit{ReLU} activation threshold, where small variations in input can determine the opening of the \textit{ReLU gate}.}
    \label{fig:neurons}
\end{figure}

\clearpage

\section{Recurrent weight matrices and Kick-circuits}
\label{sec:circuit}
\begin{figure}[h]
    \centering
    \includegraphics[width=1.0\textwidth]{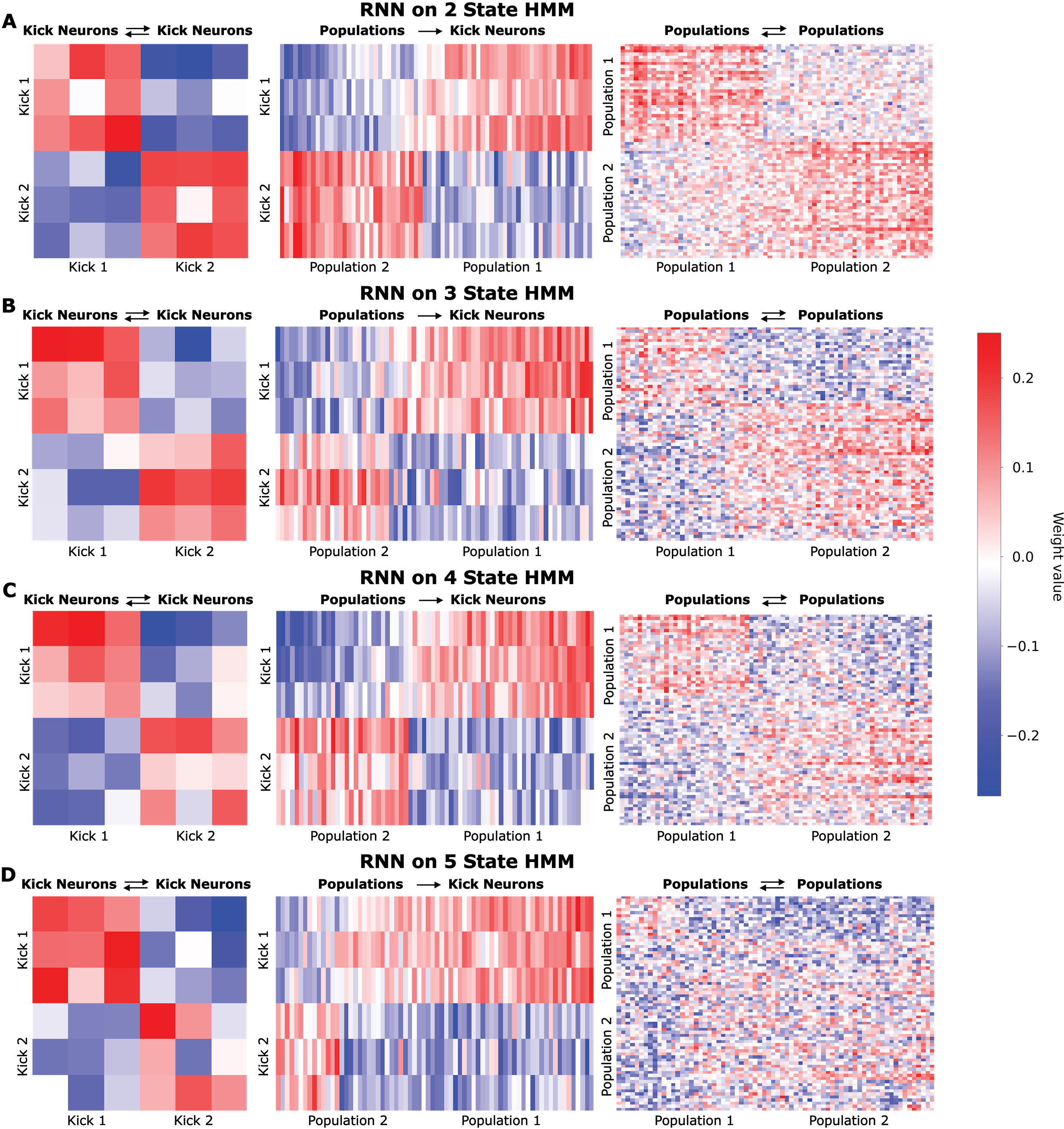}
    \caption{\textbf{Recurrent weight matrices and \textit{Kick-circuits} for linear-chain HMMs.} Each row (\textbf{A–D}) corresponds to an RNN trained to reproduce the emission statistics of a \textit{linear-chain} HMM with \(M \in \{2, 3, 4, 5\}\) latent states. \textbf{Left}: Sub-matrices of the recurrent weights $W_{hh}$ restricted to the identified \textit{kick-neurons}. Within-triplet connections are predominantly excitatory (red), while cross-triplet connections are inhibitory (blue), indicating mutual excitation and reciprocal inhibition between opposing kick groups. \textbf{Center}: Weights from the \textit{noise-integrating populations} to the \textit{kick-neurons} (sorted). Each population excites one kick-neuron triplet while inhibiting the other, implementing selective gating of transition direction. \textbf{Right}: Recurrent weights within and across the two \textit{noise-integrating populations}, showing strong within-population excitation and cross-population inhibition. Across all architectures, this structured connectivity forms a \textit{kick-circuit}: two self-exciting, mutually inhibiting loops that project to opposing \textit{kick-neurons}, enabling noise-driven, direction-selective transitions between cluster regions.}
    \label{fig:circuit1}
\end{figure}

\clearpage

\begin{figure}[h]
    \centering
    \includegraphics[width=1.0\textwidth]{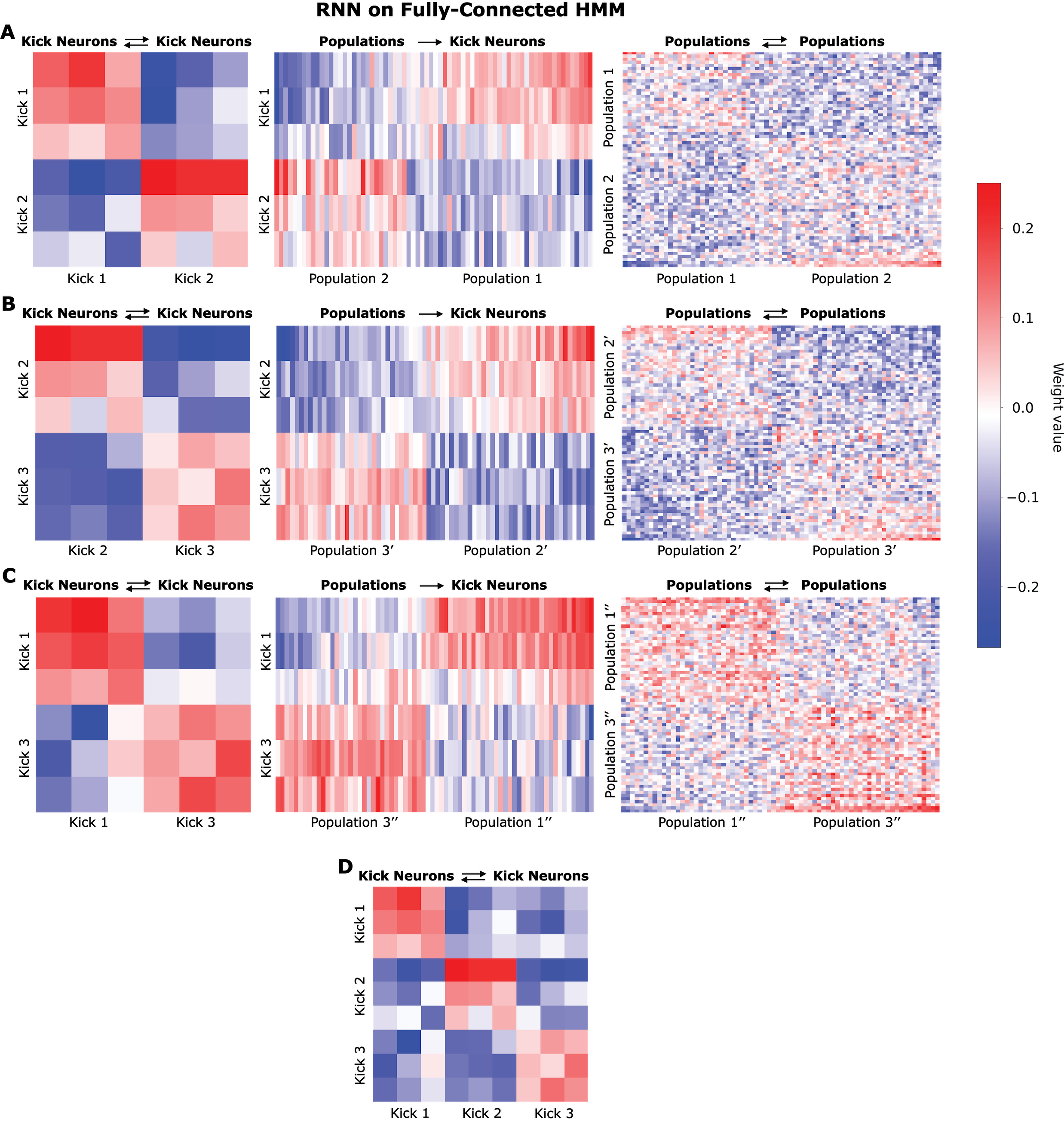}
    \caption{\textbf{Recurrent weight matrices and \textit{Kick-circuits} for fully-connected HMMs.} Each row (\textbf{A–C}) corresponds to one of the three principal subspaces identified in the RNN trained to reproduce the emission statistics of a \textit{fully-connected} HMM. Within each subspace, we isolate the \textit{kick-neurons} and associated \textit{noise-integrating populations} that together implement a self-contained instance of the dynamical primitive described for the \textit{linear-chain} architectures. \textbf{Left}: sub-matrices of the recurrent weights $W_{hh}$ restricted to the identified \textit{kick-neurons}. Within-group connections are predominantly excitatory (red), while cross-group connections are inhibitory (blue), indicating mutual excitation and reciprocal inhibition between opposing kick groups. \textbf{Center}: weights from the two \textit{noise-integrating populations} to the corresponding \textit{kick-neurons} (sorted). Each population excites one kick group while inhibiting the other, implementing selective gating of transition direction. \textbf{Right}: recurrent weights within and across \textit{noise-integrating populations}, showing strong within-population excitation and cross-population inhibition, mirroring the architecture observed in \textit{linear-chain} models. Together, these subspaces (\textbf{A–C}) represent three instances of the same \textit{kick-circuit} motif, confirming that the RNN decomposes the fully-connected HMM into compositional dynamical primitives that each generate noise-driven, direction-selective transitions. The notation $\mathrm{population}’$ (panel \textbf{B}) and $\mathrm{population}’'$ (panel \textbf{C}) indicates that population groups are re-identified independently for each subspace, whereas the \textit{kick-groups} remain fixed across panels. \textbf{D}: at the higher compositional level, the recurrent weights among the three \textit{kick-groups} show the same organization — self-excitation within each group and cross-inhibition between groups — revealing that the same circuit principle recurs hierarchically across levels, from single-neuron motifs to multi-orbit compositions.}
    \label{fig:circuit2}
\end{figure}

\clearpage

\section{Alignment of Output Dimensions with Orbits Plane}
\label{sec:alignment}
\begin{figure}[h]
    \centering
    \includegraphics[width=0.7\textwidth]{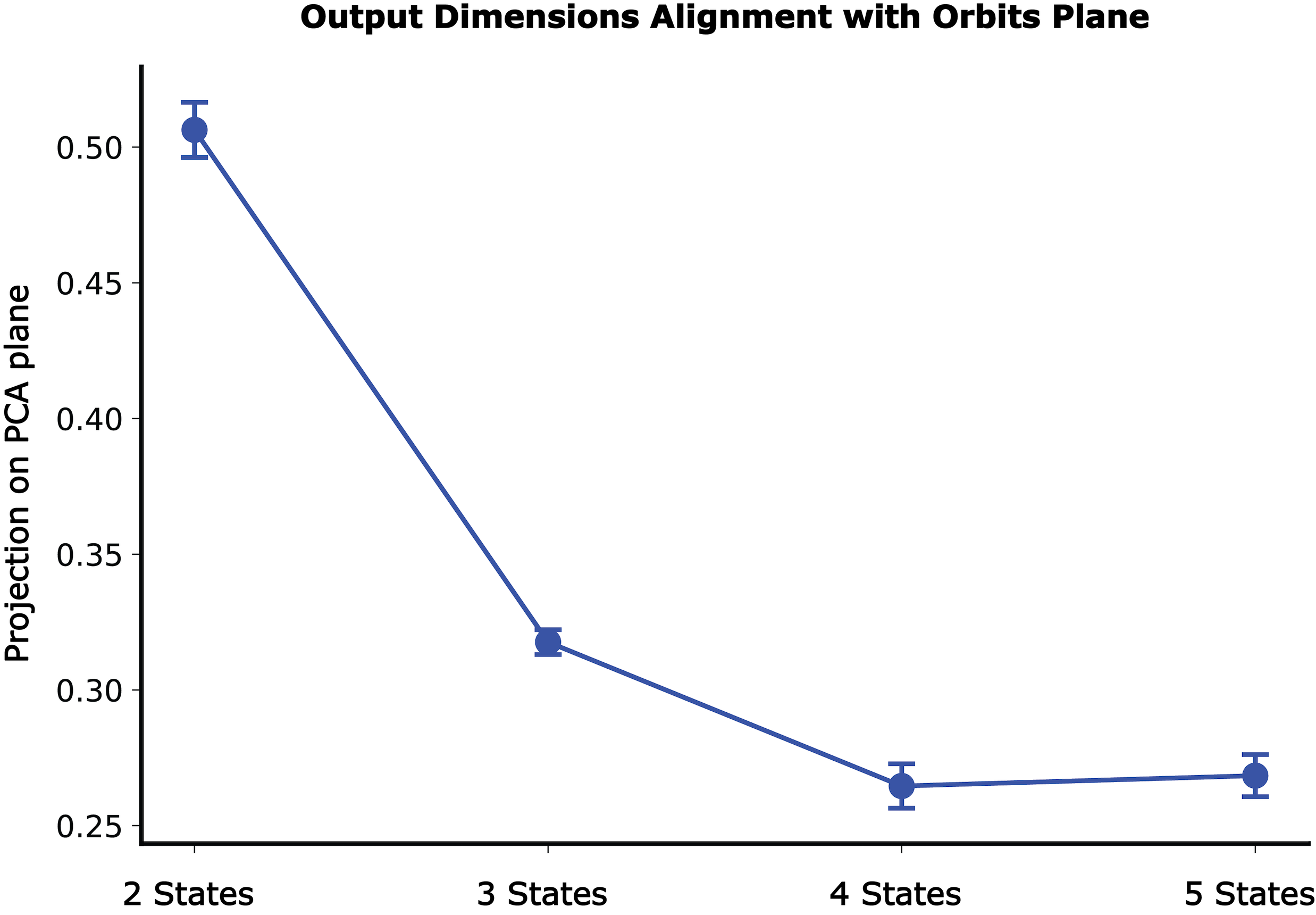}
    \caption{\textbf{Alignment of output dimensions with orbit plane.} Mean $\pm$ s.d. projection of the three output readout axes onto the principal plane spanned by the orbital dynamics, computed across three independently trained RNNs \((|h|=150,\,d=100)\) for each \textit{linear-chain} HMM with \(M \in \{2, 3, 4, 5\}\) latent states. As the number of latent states increases, the alignment between the readout axes and the orbit plane systematically decreases, indicating that the networks progressively encode output probabilities through subtler modulations within a shared low-dimensional dynamical manifold rather than by forming additional slow regions or distinct attractors. This geometric reorientation provides a mechanistic account of how RNNs approximate increasingly fine latent discretizations — transitioning from discrete to quasi-continuous regimes — by reusing the same dynamical primitive while generating smoother, less skewed output distributions.}
    \label{fig:alignment}
\end{figure}

\clearpage
\section{Limitations and Broader Impact}
\subsection{Discussion of Limitations}
\label{subsec:limitations}
This work focuses on RNNs trained to emulate relatively simple HMM structures. Despite the diversity of HMM families examined (\textit{linear-chain, fully-connected} and \textit{cyclic}), the target models share symmetric transition graphs and emission probabilities constrained to two or three output classes. While this design allows for clear interpretation and mechanistic analysis, natural behaviors are often governed by richer latent structures featuring asymmetric transitions and more diverse emission profiles. Extending the present framework to capture these more complex dynamics remains an important direction for future work. Nevertheless, preliminary analyses on naturalistic animal behavior suggest that the core dynamical primitive and its underlying mechanisms identified here generalize beyond the simplified settings considered, providing a potential foundation for modeling spontaneous behavioral sequences in biological systems.

\subsection{Broader Impact}
\label{subsec:broader-impact}
This work aims to advance our understanding of how Recurrent Neural Networks (RNNs) can represent discrete, stochastic latent structure through continuous dynamics. By reverse-engineering trained RNNs, we uncover a mechanistic account of how such representations can emerge in the absence of explicit structural constraints. This contributes to a growing body of work exploring how neural architectures can serve as data-driven models of behavior.

A key implication is that RNNs may offer a powerful alternative to traditional models like Hidden Markov Models (HMMs) for inferring latent structure in naturalistic behaviors. Unlike HMMs, which impose strong assumptions about the topology of the latent state space, RNNs allow for both discrete and continuous structures to emerge directly from the data. This flexibility has the potential to yield more biologically plausible representations of the computations underlying behavior.

More broadly, the dynamical mechanisms we identify — particularly how structured connectivity and noise interactions give rise to stochastic, quasi-periodic transitions — may inspire novel hypotheses about the neural representations of spontaneous, sequential behaviors in biological systems.

We do not foresee any direct negative societal impacts arising from this work.

\end{document}